\documentclass[10pt,twocolumn,letterpaper]{article}

\usepackage{iccv}
\usepackage{times}
\usepackage{epsfig}
\usepackage{graphicx}
\usepackage{amsmath}
\usepackage{amssymb}

\usepackage{xcolor}
\usepackage{bm} 
\usepackage{mathtools} 
\usepackage{subcaption}

\usepackage[pagebackref=true,breaklinks=true,letterpaper=true,colorlinks,bookmarks=false]{hyperref}

\iccvfinalcopy 

\def\iccvPaperID{8200} 

\ificcvfinal\fi

\DeclareMathOperator{\tr}{tr}

\newcommand{\halfline}{\vspace{0.5\baselineskip}}

\newcommand{\x}{\mathbf{x}}

\begin{document}

\title{Bayesian Deep Basis Fitting for Depth Completion with Uncertainty
\thanks{
This work was supported in part by C-BRIC, one of six centers in JUMP, a Semiconductor Research Corporation (SRC) program sponsored by DARPA.
This work was supported in part by the Semiconductor Research Corporation (SRC) and DARPA.
We gratefully acknowledge the support of NVIDIA Corporation with the donation of the DGX Station/Jetson TX2/etc. used for this research.
}
}

\author{Chao Qu \hspace{2cm} Wenxin Liu \hspace{2cm} Camillo J. Taylor \\
University of Pennsylvania\\
{\tt\small \{quchao, wenxinl, cjtaylor\}@seas.upenn.edu}
}

\maketitle
\ificcvfinal\thispagestyle{empty}\fi

\begin{abstract}
In this work we investigate the problem of uncertainty estimation 
for image-guided depth completion.
We extend Deep Basis Fitting (DBF) \cite{Qu2020DepthCV} for depth completion 
within a Bayesian evidence framework to provide calibrated per-pixel variance. 
The DBF approach frames the depth completion problem in terms of 
a network that produces a set of low-dimensional depth bases 
and a differentiable least squares fitting module 
that computes the basis weights using the sparse depths. 
By adopting a Bayesian treatment, 
our Bayesian Deep Basis Fitting (BDBF) approach is able to 
1) predict high-quality uncertainty estimates and 
2) enable depth completion with few or no sparse measurements.
We conduct controlled experiments to compare BDBF against 
commonly used techniques for uncertainty estimation 
under various scenarios.
Results show that our method produces 
better uncertainty estimates with accurate depth prediction.
\end{abstract} 
\section{Introduction}

\begin{figure}[t]
\centering
\includegraphics[width=\linewidth]{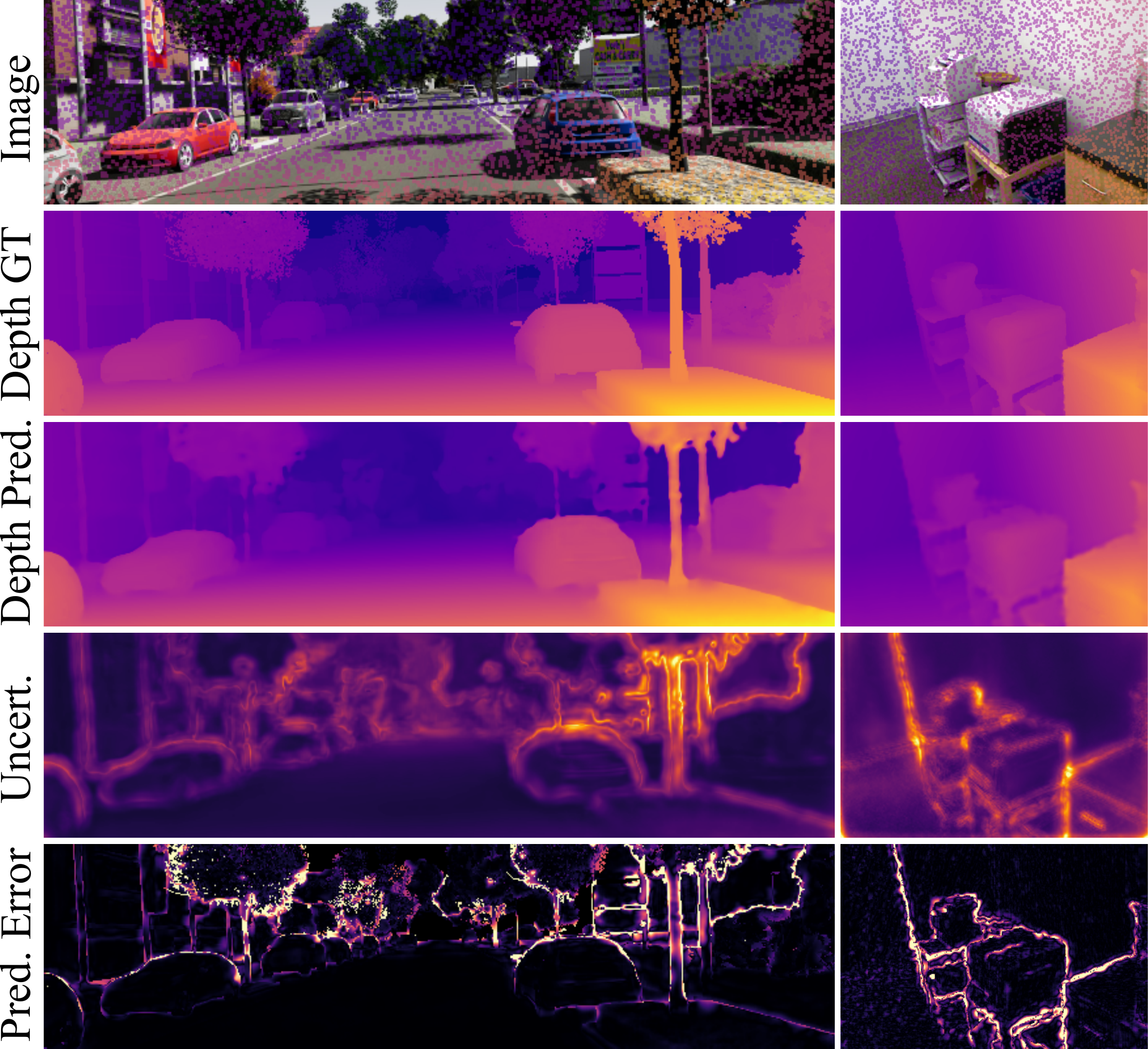}
\caption{
Qualitative results of our method, 
Bayesian Deep Basis Fitting (BDBF), 
which outputs uncertainty estimates 
with depth completion.
}
\label{fig:bdbf_plots}
\end{figure}

As we seek to incorporate learned modules in safety critical applications such as
autonomous driving, reliable uncertainty estimation becomes as critical as prediction accuracy ~\cite{Snderhauf2018TheLA}.
Depth completion is one such task where well-calibrated uncertainty estimates 
can help to enable robust machine perception.
Deep Convolutional Neural Networks (CNNs) are commonly used 
to solve structured regression problems like depth prediction 
due to their strong expressive power and inductive bias~\cite{Eigen2014DepthMP}.
However, in its native form, a CNN only produces a point estimate, 
which offers few insights into whether or where its output should be trusted.
Many probabilistic deep learning methods have been proposed to address 
this issue~\cite{MacKay1992APB,Gal2016UncertaintyID},
but they often fail to output calibrated uncertainty~\cite{Guo2017OnCO}
or become susceptible to distributional shift~\cite{Ovadia2019CanYT}.
Moreover, these methods can be expensive to compute due to
the need for test time sampling~\cite{Gal2016DropoutAA} 
or inference over multiple models~\cite{Lakshminarayanan2017SimpleAS}.

In this work, we propose a method for depth completion with uncertainty estimation 
that avoids the above limitations.
Our approach builds on the idea of Deep Basis Fitting (DBF)~\cite{Qu2020DepthCV}.
DBF replaces the last layer of a depth completion network with a set of \textit{data-dependent} weights.
These weights are computed by a differentiable least squares fitting module
between the penultimate features and the sparse depths.
The network can also be seen as an adaptive basis function 
which explicitly models scene structure on a low-dimensional manifold~\cite{Bloesch2018CodeSLAML, Tang2019BANetDB}.
It can be used as a replacement to the final layer 
(with no change to the rest of the network or training scheme),
which greatly improves depth completion performance.

We extend DBF by formulating it within a Bayesian evidence framework \cite{Bishop2006PatternRA}.
This is done by placing a prior distribution on the DBF weights and marginalizing it out during inference.
Such last-layer probabilistic approach have been shown to be 
reasonable approximations to full Bayesian Neural Networks~\cite{Kristiadi2020BeingBE},
while providing the advantage of tractable inference \cite{Ober2019BenchmarkingTN}.
This is conceptually similar to Neural Linear Models (NLMs) \cite{Snoek2015ScalableBO}
with the notable distinction that we perform Bayesian linear regression 
on each image as opposed to the entire dataset.

A Bayesian treatment also enables depth completion with highly sparse data.
In DBF, when the number of sparse depths falls below the dimension of the bases,
the underlying linear system becomes under-determined.
We show that by learning a shared prior across images, 
our method is able to handle any number of sparse depth measurements.

We name our approach Bayesian Deep Basis Fitting (BDBF) and summarize 
its advantages:
1) It can be used as a drop-in replacement to the final layer 
of many depth completion networks and outputs uncertainty estimates
(in the form of per-pixel variance).
2) Compared to other uncertainty estimation techniques, 
it produces higher quality uncertainty with one training session, 
one saved model and one forward pass, without needing extra parameters or modifications to the loss function.
3) It can handle any sparsity level, 
with performance degrading gracefully towards a pure monocular method 
when the number of depth measurements goes to zero.

\section{Related Work}

\subsection{Uncertainty Estimation for Neural Networks}

We start by reviewing uncertainty estimation techniques for neural networks.
There are two types of uncertainty that one could model: 
\textit{epistemic (model)},  which describe the uncertainty in the model 
and \textit{aleatoric (data)}, which reflects the inherent noise in the data~\cite{Kiureghian2009AleatoryOE}. 
Modeling uncertainty in neural networks can be achieved by 
placing probabilistic distributions on network weights. 
Such networks are called Bayesian Neural Networks (BNN)~\cite{MacKay1992APB}. 
Direct inference in BNNs is intractable for continuous variables, 
and different approximation techniques have been explored~\cite{MacKay1992APB, jordan1999introduction,graves2011practical,blundell2015weight, neal1995, minka2001family,jylanki2014expectation}. 
However, they don't scale well to large datasets and complex models,
and are thus impractical for current vision tasks.

Gal \etal~\cite{Gal2016DropoutAA} proposed the use of \textit{dropout} 
as an approximate variational inference method for BNNs.
However, their method requires multiple forward passes to obtain Monte Carlo model estimates at test time.
Another research direction is \textit{assumed density filtering} (ADF)~\cite{opper1998bayesian} 
which can be viewed as a single Expectation Propagation pass. 
Gast \etal~\cite{Gast2018LightweightPD} chose to propagate activation uncertainties 
without probabilistic weights in a lightweight manner, 
which requires modifying the layer operations based on moment matching.

\textit{Predictive} methods directly output mean and variance 
of some parametric distribution by minimizing the 
negative-log likelihood (NLL) loss~\cite{Nix1994EstimatingTM}.
They only require small changes to the original network by adding a variance prediction head.
This simplicity makes it a popular choice among recent works~\cite{Kendall2017WhatUD, Liu2020TLIOTL}. 

\textit{Ensemble} methods either train multiple models independently with 
different initializations (\textit{bootstrap})~\cite{Lakshminarayanan2017SimpleAS} 
or save several copies of weights at different stages during training 
(\textit{snapshot})~\cite{Huang2017SnapshotET}. 
These methods only model epistemic uncertainty 
but can be combined with a predictive one to model data uncertainty.
They achieve good performances in various experimental
settings~\cite{Gustafsson2020EvaluatingSB,Poggi2020OnTU,Ilg2018UncertaintyEA}, 
but still need multiple inference passes at test time,
which makes them less suitable for resource constrained platforms. 

Table~\ref{tab:method_comp} summarizes the aforementioned approaches 
and highlights the difference compared to ours.
In Sec.~\ref{sec:eval_base} we describe in detail the methods that we evaluate against.

\begin{table}
\centering\footnotesize
\begin{tabular}{|l|r|r|r|c|c|}
\hline
Method & \#T & \#M & \#F & Alea. & Epis. \\ \hline
Predictive \cite{Nix1994EstimatingTM, Kendall2017WhatUD} & 1   & 1  & 1  & \checkmark &    \\ \hline
Dropout (Predictive) \cite{Gal2016DropoutAA,Kendall2017WhatUD}   & 1 & 1 & K & (\checkmark) & \checkmark  \\
Snapshot (Predictive) \cite{Huang2017SnapshotET} & 1 & K  & K  & (\checkmark) & \checkmark \\
Bootstrap (Predictive) \cite{Lakshminarayanan2017SimpleAS} & K & K & K & (\checkmark) & \checkmark \\ \hline
\textbf{Proposed (BDBF)} & 1 & 1 & 1  & \checkmark & \checkmark \\ \hline
\end{tabular}
\caption{
Comparison of different uncertainty estimation techniques. 
The first three columns represent number of training sessions (T), copies of model saved (M), and number of forward passes (F) at test time respectively. 
Last two columns indicate whether a method estimates data or model uncertainties. 
Dropout, snapshot and bootstrap ensemble can all be combined with a predictive approach to model data uncertainty.
BDBF has the same complexity as predictive methods.
}
\label{tab:method_comp}
\end{table}

\subsection{Uncertainty Estimation in Depth Completion}

Great progress has been made in the past few years on depth completion ranging from 
high-density completion for RGB-D/ToF cameras~\cite{Liu2013GuidedDE, Ferstl2013ImageGD, Xue2017DepthII, Zhang2018DeepDC},
to mid-density completion from LiDAR sensors~\cite{Uhrig2017SparsityIC, Ma2018SparsetoDenseDP, Chodosh2018DeepCC, Tang2019LearningGC, Cheng2020CSPNLC, Chen2019LearningJ2, Zhao2020AdaptiveCM, Li2020AMG, Xu2019DepthCF, Yang2019DenseDP}.
Recently, there has also been rising interests in low-density completion 
from map points generated by Visual-SLAM or Visual-Inertial Odometry~\cite{Wang2018PlugandPlayID, Wong2020UnsupervisedDC,Sartipi2020DeepDE, Zuo2020CodeVIOVO}.
Unlike systems that are designed for a particular sparsity or sensing modality,
our proposed method can be seen as a general component for depth completion 
similar to DBF~\cite{Qu2020DepthCV}.

A complete review of depth completion literature is out of the scope of this work, 
we instead focus on methods that also \textbf{estimate uncertainty}.
Gansbeke \etal~\cite{Gansbeke2019SparseAN} predict depth and confidence weights 
for both color and depth branch and fuse them based on the confidence maps.
Qiu \etal~\cite{Qiu2018DeepLiDARDS} adopt a similar strategy, 
but additionally guide the depth branch via surface normal prediction.
Xu \etal~\cite{Xu2019DepthCF} use a shared encoder and multiple decoders to predict
surface normal, coarse depth and confidence, 
then use an anisotropic diffusion process to produce refined depth.
Park \etal~\cite{Park2020NonLocalSP} instead use a single encoder-decoder network 
to predict initial depth, affinities and confidence
before applying a non-local spatial propagation to produce the final depth.
Note that the uncertainties produced by the above methods are not calibrated 
and are only used internally. 
Therefore, they are not readily useful to downstream tasks that require probabilistic reasoning.
This type of uncertainty estimation can also be seen as 
a simplified version of the \textit{predictive} method in~\cite{Nix1994EstimatingTM} without the NLL loss.

Few works have tried to evaluate the quality of depth completion uncertainty.
Eldesokey \etal~\cite{Eldesokey2020UncertaintyAwareCF} present 
a probabilistic normalized convolution~\cite{Eldesokey2018PropagatingCT} 
that estimates confidence of both input sparse depths and output dense prediction,
for unguided depth completion.
Gustafsson \etal~\cite{Gustafsson2020EvaluatingSB} compared 
several uncertainty estimation methods applied to depth completion 
in the same spirit as~\cite{Poggi2020OnTU}. 
We follow their approach and provide a systematic comparison of our proposed method 
against the best performing schemes from~\cite{Gustafsson2020EvaluatingSB, Poggi2020OnTU}
and demonstrate superior performance and efficiency across a range of datasets.
  
\section{Method}
\subsection{Problem Formulation}
\begin{figure}
\centering
\includegraphics[width=\linewidth]{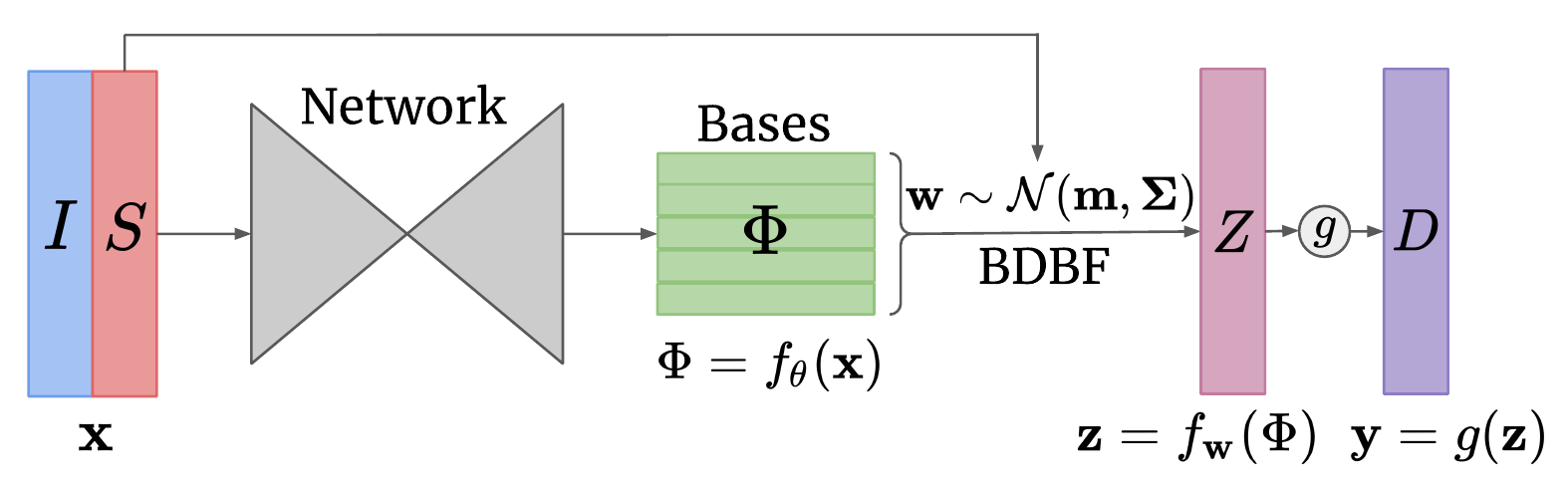}
\caption{An overview of BDBF for depth completion.
The input to the network is an RGB image $I$ and optionally a sparse depth map $S$.
The network produces a $M$-dimensional depth bases $\bm{\Phi}$, 
which has the same spatial resolution as $I$.
BDBF then solves for the weights $\mathbf{w}$ given the sparse depth at valid pixel locations.
$\mathbf{w}$ can then be used to reduce the bases into a single channel latent prediction $Z$,
before going through the activation $g$ to produce depth $D$.
}
\label{fig:method/overview}
\end{figure}

Let $\mathcal{D} = \{(\mathbf{x}_n, \mathbf{y}_n)\}_{n=1}^{N_\mathcal{D}}$ 
be a dataset containing $N_\mathcal{D}$ samples. 
We wish to learn a neural network $f$ that maps $\mathbf{x}$ to $\mathbf{y}$. 
In depth completion, 
the input $\mathbf{x}$ is usually 
an image and sparse depth pair $(I, S)$,
and the output $\mathbf{y}$ is the predicted depth map $D$. 
We refer to $f_\theta$ as the basis network
and its output $\Phi$ a set of \textit{depth bases}~\cite{Qu2020DepthCV}.
$\Phi$ is then reduced by a \textit{linear} layer $f_\mathbf{w}$ to $\mathbf{z}$,
which is then mapped to positive depth values via 
a \textit{nonlinear} activation function $g$.
\begin{align}\label{eq:prelim_map}
\mathbf{y}
= f (\mathbf{x}) 
= g \circ f_\mathbf{w} \circ f_\theta(\mathbf{x}) 
= g \circ f_\mathbf{w} (\Phi)
= g(\mathbf{z})
\end{align}
With a slight abuse of notation, 
we call $\mathbf{z}$ the \textit{latent} variable
and choose $g$ to be the exponential function~\cite{Eigen2014DepthMP},
so $\mathbf{z}$ effectively corresponds to log depth.
An overview of our method is shown in Figure~\ref{fig:method/overview}.

\subsection{Bayesian Deep Basis Fitting}
\label{sec:method_bdbf}
We choose to model the distribution of each pixel in the latent space $\mathbf{z}$ 
rather than in the target space $\mathbf{y}$, since depth is strictly positive
and may span several orders of magnitude \cite{Snelson2003WarpedGP}.
Assuming Gaussian noise in the latent space, we define our model to be
\begin{align}
z_i = \mathbf{w}^\top \bm{\phi}_i + \epsilon_i, 
\quad  
\epsilon_i \sim \mathcal{N}(0, \beta^{-1})
\end{align}
where $\bm{\phi}_i$ denotes the basis entries corresponding to the latent pixel value $z_i$ 
and $\beta \in \mathbb{R}$ is a precision parameter 
that corresponds to the inverse variance of the noise.
Assuming that the errors at each pixel are independent, the likelihood function is
\begin{align}
p(\mathbf{z} | \mathbf{x}, \mathbf{w})
&= \mathcal{N}(\mathbf{z} | \bm{\Phi} \mathbf{w}, \beta^{-1} \mathbf{I})
\end{align}
here $\bm{\Phi}$ is the $N\times M$ \textit{design matrix}
with $N$ the number of sparse depths and $M$ the dimension of $\mathbf{w}$.
It is assembled by extracting the basis entries at the pixel locations specified by $S$. 

Given a suitable prior on the last-layer weights 
$p(\mathbf{w}) = \mathcal{N}(\mathbf{m}_0, \alpha^{-1}\bm{\Sigma}_0)$, 
where $\alpha\in \mathbb{R}$ is a precision parameter to scale the covariance
$\bm{\Sigma}_0$,
the posterior distribution of $\mathbf{w}$ can be computed analytically 
following Bayes' rule~\cite{Bishop2006PatternRA}:
\begin{align}
p(\mathbf{w}|\mathbf{x}, \mathbf{z}) 
&= \mathcal{N}(\mathbf{m}, \bm{\Sigma}) \\
&\propto \mathcal{N} (\mathbf{w} | \mathbf{m}_0, \alpha^{-1}\bm{\Sigma}_0) 
\cdot \mathcal{N}(\mathbf{z} | \bm{\Phi} \mathbf{w}, \beta^{-1} \mathbf{I})
\end{align}
where the mean and covariance are given by 
\begin{align}
\mathbf{m} 
&= \bm{\Sigma}(\alpha\bm{\Sigma}_0^{-1} \mathbf{m}_0 + \beta \bm{\Phi}^\top \mathbf{z})
\label{eq:posteriror_mean}
\\
\bm{\Sigma}
&= (\alpha\bm{\Sigma}_0^{-1} + \beta \bm{\Phi}^\top \bm{\Phi} )^{-1}
\label{eq:posterior_cov}
\end{align}
The latent predictive distribution for a pixel at test time is
\begin{align}
p(z_*|\mathbf{x}, \mathbf{z}) 
&= \int p(z_*|\mathbf{w}) p(\mathbf{w}|\mathbf{x},\mathbf{z})d{\mathbf{w}} \\
&= \mathcal{N}(z_* | \mathbf{m}^\top \bm{\phi}_*, \bm{\phi}_*^\top  \bm{\Sigma} \bm{\phi}_*)
\label{eq:pred_dist}
\end{align}
The Gaussian assumption is made solely for the purpose of tractable inference.
In practice, the shape of the predictive distribution depends heavily on the loss function.
Since we use L1 loss for training, 
we employ a Laplace distribution as its parametric form for evaluating uncertainty~\cite{Ilg2018UncertaintyEA,Bloesch2018CodeSLAML}.
Additionally, a robust norm like Huber~\cite{Huber1964RobustEO} 
can be applied if outliers are present in the target~\cite{Triggs1999BundleA} .
\subsection{Training}
\noindent
\textbf{Loss Function.}
The standard way of learning such models is by 
maximizing the marginal likelihood function 
with respect to the parameters $\theta$ of the basis function $f_\theta$, 
\begin{align}
\begin{split}\label{eq:marg_like}
\log p(\mathbf{z}| \x,\alpha, \beta) 
&= \frac{1}{2} (
N \ln \beta + M \ln \alpha - N \ln 2\pi  \\ 
& \qquad  - E(\mathbf{m}) + \ln |\bm{\Sigma}| - \ln |\bm{\Sigma}_0|)
\end{split} \\
E(\mathbf{w}) 
&= \beta \|\mathbf{z} - \bm{\Phi}\mathbf{w}\|^2
+ \alpha \|\mathbf{w} - \mathbf{m}_0\|^2_{\bm{\Sigma}_0}
\end{align}
where $\|\mathbf{v}\|_\mathbf{A} = \mathbf{v}^\top \mathbf{A}^{-1} \mathbf{v}$ is the Mahalanobis norm. 
This is known in literature as \textit{type 2 maximum likelihood} \cite{Bishop2006PatternRA}.

Directly maximizing~\eqref{eq:marg_like} has two practical issues. 
First, one needs to estimate the hyperparameters $\alpha$ and $\beta$, 
which adds large overhead during training.
Second, gradients need to be back-propagated through 
costly operations like matrix inversion and determinant.
Together, they pose challenges to the training phase 
and often produce empirically similar results to 
a point estimate~\cite{Snoek2015ScalableBO}.
We avoid these issues by assuming sufficient sparse points in training ($N \gg M$). 
This renders the linear system over-determined,
which allows us to treat the prior $p(\mathbf{w})$ as infinitely broad.
The solution in~\eqref{eq:posteriror_mean} therefore reduces to a maximum likelihood (ML) one 
which can be computed efficiently in one pass~\cite{Bishop2006PatternRA}.
\begin{align}\label{eq:w_ml}
\mathbf{w}_\mathrm{ML} 
&= (\bm{\Phi}^\top \bm{\Phi})^{-1} \bm{\Phi}^\top \mathbf{z} 
\\
\beta_\mathrm{ML}^{-1} 
&= \frac{1}{N} \| \mathbf{z} - \bm{\Phi} \mathbf{w}_\mathrm{ML} \|^2
\end{align}

The predicted mean and variance of $z_*$ at training time
are given by the following according to~\eqref{eq:pred_dist}
\begin{align}\label{eq:pred_ml}
\mu 
= \mathbf{w}_\mathrm{ML}^\top \bm{\phi},\quad
\sigma^2
= \beta_\mathrm{ML}^{-1} \bm{\phi}^\top (\bm{\Phi}^\top \bm{\Phi})^{-1} \bm{\phi}
\end{align}

Given the above results, we can minimize the Negative Log-Likelihood Loss (NLL) 
assuming a Laplace distribution~\cite{Ilg2018UncertaintyEA}, 
which is defined for a single pixel as 
\begin{align}\label{eq:loss_nll}
-\log p(z | \mu, b) 
\propto \frac{|\mu - z|}{b} + \log{b}, \quad 2b^2=\sigma^2
\end{align}
However, this would still involve back-propagation through a costly matrix inversion as in~\eqref{eq:w_ml}.
We find in our experiments that this sometimes causes numerical instability during training 
and incurs a visible decrease in prediction accuracy.
Therefore, we opt to minimize directly the L1 loss for supervised learning.
This allows our network to be trained in its original form
without suffering from the performance drop caused by the NLL loss~\cite{Loquercio2020AGF}.

\halfline
\noindent
\textbf{Uncertainty Calibration.}
Not using the likelihood loss comes with the risk of over-confidence in uncertainty estimation,
since there is no explicit penalty on the variance prediction.
As the number of parameters in $\theta$ is usually on the order of millions, 
the noise variance $\beta^{-1}$ will be pushed towards zero~\cite{Ober2019BenchmarkingTN}.
One solution is to regularize $\theta$ using an L2 regularization term in the optimizer.
This introduces an extra hyperparameter to tune: 
a small regularization would not prevent overfitting, 
while a large one will render the feature bases inexpressive~\cite{Thakur2020LearnedU}. 
We notice empirically that the amount of overconfidence in our method 
is consistent during training and validation.
Therefore, we take a pragmatic approach and propose to solve 
this problem in terms of estimator consistency \cite{Bailey2006ConsistencyOT}, 
measured by normalized estimation error squared (NEES).
For a Laplace distribution, NEES is defined as 
$\varepsilon = (\mu - z)^2/b^2$.
We record the average NEES $\bar{\varepsilon}$ at training time for the last epoch,
and use it to scale the variance accordingly during inference with
$\bar{\sigma}^2 = \bar{\varepsilon} \sigma^2$,
which attempts to make the final prediction consistent.
Note that the scaling factor is computed entirely at training 
without any additional data and NEES is not used as a loss function.

\halfline
\noindent
\textbf{Shared Prior.}
Although the prior $p(\mathbf{w})$ is not used in training, we still need it for inference.
Rather than estimating a different prior for each image, we make another simplifying assumption that there exists a shared prior for the entire dataset.
This aligns with our observation from experiments that $p(\mathbf{w})$ shows a relatively sharp peak.
Given our training strategy, we adopt a frequentist approach and collect all ML solutions of weights $\mathbf{w}_\mathrm{ML}$ within one training epoch.
The mean, $\mathbf{m}_0$, and covariance, $\bm{\Sigma}_0$, can then be computed from this set.
Having a shared prior enables robust depth completion from a few sparse depth measurements.
\subsection{Inference}
Inference follows the standard evidence framework~\cite{Bishop2006PatternRA}.
We use EM~\cite{Dempster1977MaximumLF} to estimate the hyperparameters $\alpha$ and $\beta$.
The re-estimation equations are obtained by maximizing 
the expected complete-data log likelihood with respect to $\alpha$, $\beta$
\begin{align}\label{eq:em_reest}
\alpha^{-1} 
&= \frac{1}{M} \left(\|\mathbf{m} - \mathbf{m}_0 \|^2_{\bm{\Sigma_0}} 
+ \tr(\bm{\Sigma}_0^{-1}\bm{\Sigma} ) \right)\\
\beta^{-1} 
&= \frac{1}{N}\left( \|\mathbf{z} - \bm{\Phi} \mathbf{m}\|^2 
+ \tr(\bm{\Phi}^\top \bm{\Phi} \bm{\Sigma}) \right)
\end{align}
where $\tr(\cdot)$ is the matrix trace operator.
The re-estimated $\alpha$ and $\beta$ are then plugged back into \eqref{eq:posteriror_mean} and \eqref{eq:posterior_cov}
to recompute $\mathbf{m}$ and $\bm{\Sigma}$.
We initialize this process empirically with $\alpha=1$ and $\beta = \sqrt{N}$ 
and set the maximum number of iterations to 8.
In practice we reach convergence within 2 to 3 iterations when $N \gg M$, 
thus incurring only a small computation overhead.
In the extreme case when $N\rightarrow 0$, 
we rely on the shared prior alone for a pure monocular prediction.
\begin{align}
\mu = \mathbf{m}_0^\top \bm{\phi},
\quad
\sigma^2 = \bm{\phi}^\top \bm{\Sigma}_0 \bm{\phi}
\end{align}
\begin{figure*}
\centering
\includegraphics[width=\linewidth]{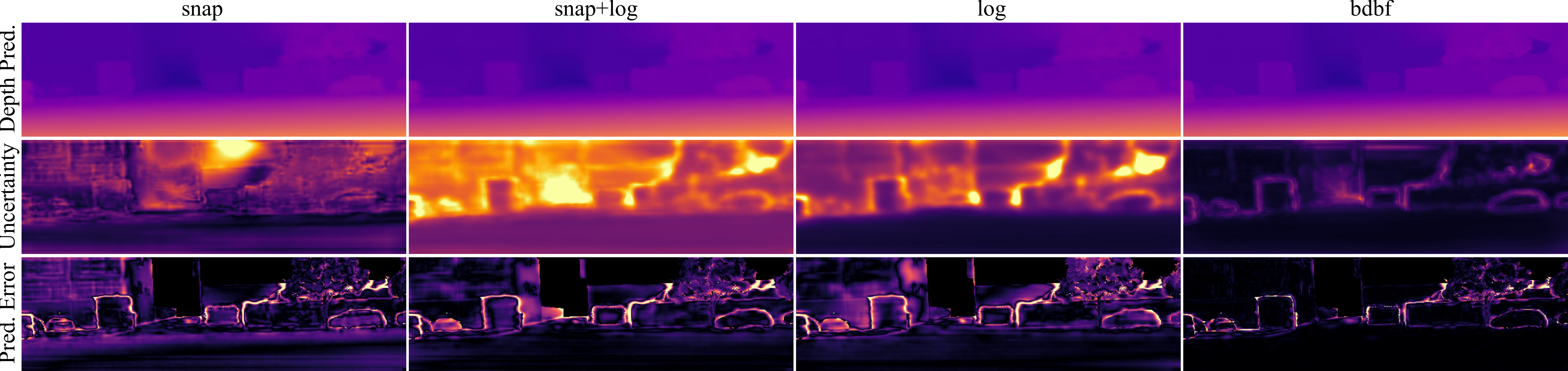}
\caption{
Qualitative results of all methods trained and test with 5\% sparsity on VKITTI2. 
}
\label{fig:mid_vk2_qual}
\end{figure*}

\section{Evaluation}
In this section, we thoroughly evaluate our proposed method (\textit{bdbf}).
We show with extensive experimental results on three different datasets 
and various sparsity settings that our method outperforms baseline approaches 
in uncertainty estimates with accurate depth prediction, 
and remains resilient to sparsity change and domain shift.
\subsection{Baselines}\label{sec:eval_base}
We describe three baselines for uncertainty estimation that we compare to,
which are shown to have strong performance~\cite{Poggi2020OnTU,Gustafsson2020EvaluatingSB}
and can be evaluated under controlled settings.
As discussed in Section~\ref{sec:method_bdbf},
all methods output mean and variance of the latent prediction before $g(\cdot)$,
and are trained with either L1 loss or its NLL variant~\eqref{eq:loss_nll}.

For empirical methods, we choose Snapshot Ensemble \cite{Huang2017SnapshotET}
(\textit{snap}) which can be completed in one training session 
such that all methods have the same training budget.
The mean and variance are computed using $K$ snapshots.
\begin{align}\label{eq:prelim_emp}
\mu      = \frac{1}{K} \sum_{i=1}^K \mu_i, \quad 
\sigma^2 = \frac{1}{K} \sum_{i=1}^K (\mu_i - \mu)^2
\end{align}
For predictive methods (\textit{log})~\cite{Nix1994EstimatingTM}, 
we attach a variance prediction head parallel to the depth prediction and train with the NLL loss. 
Finally, we combine the above two (\textit{snap+log})~\cite{Kendall2017WhatUD} 
to form a predictive ensemble method.
\begin{align}\label{eq:prelim_bay}
\mu      = \frac{1}{K} \sum_{i=1}^K \mu_i, \quad 
\sigma^2 = \frac{1}{K} \sum_{i=1}^K \left( (\mu_i - \mu)^2 + \sigma_i^2 \right)
\end{align}
\subsection{Datasets}

\noindent
\textbf{Virtual KITTI 2.}
The VKITTI2 dataset~\cite{cabon2020vkitti2} is an updated version
to its predecessor~\cite{gaidon2016virtual}.
We use sequences 2, 6, 18 and 20 with variations 
\textbf{clone}, \textbf{morning}, \textbf{overcast} and \textbf{sunset} 
for training and validation, and \textbf{clone} in sequence 1 for testing.
This results in 6717 training and 447 testing images.
The sparse depths are generated by randomly sampling pixels that have a depth less than 80m~\cite{Gustafsson2020EvaluatingSB}.
Ground truth depths are also capped to 80m following common evaluation protocols.
All images are downsampled by half.

\halfline
\noindent
\textbf{NYU Depth V2.}
The NYU-V2 \cite{Silberman2012IndoorSA} dataset is comprised of 
various indoor scenes recorded by an off-the-shelf RGB-D camera.
We use the 1449 densely labeled pairs of aligned RGB and depth images.
and split it into approximately 75\% training and 25\% testing. 
The same depth sampling strategy is adopted as above.
Note that we intentionally choose this small dataset (as opposed to the full) to 
evaluate uncertainty estimation under data scarcity~\cite{Thakur2020LearnedU}.

\halfline
\noindent
\textbf{KITTI Depth Completion.}
We also evaluate on the KITTI depth completion dataset \cite{Uhrig2017SparsityIC}
following its official train/val split.
For all experiments other than the official submission,
we down sample both the image and depth by half.


\subsection{Implementation Details}
\noindent
\textbf{Network architecture.}
We use the same basis network for all methods, 
which is an encoder-decoder architecture with skip connection 
similar to that in~\cite{Qu2020DepthCV}.
We use a MobileNet-V2~\cite{Snelson2003WarpedGP} pretrained on ImageNet~\cite{Krizhevsky2017ImageNetCW}.
The decoder outputs a set of \textit{multi-scale bases}~\cite{Qu2020DepthCV}, 
which are then upsampled to the input resolution 
and concatenated together to form a final 63-dimensional basis.
For baseline methods we initialize the bias of depth prediction head
with the average log depth of the dataset 
and let the variance head predict an initial variance of 1.
Our method, however, requires no initialization. 
When using sparse depths as network input, we adopt the two-stage approach from \cite{Wong2020UnsupervisedDC} 
which first scaffolds the sparse depths by interpolation 
and then fuses it with the first layer of the encoder via convolution.
Note that this depth pre-processing step is orthogonal to the uncertainty estimation techniques
and we choose this approach for its simplicity and applicability to both mid- and low-sparsity.
All networks use the same setup unless otherwise stated.

\halfline
\noindent
\textbf{Training parameters.}
For training we use the Adam optimizer \cite{Kingma2015AdamAM} 
with an initial learning rate of 2e-4 and reduce it by half every 5 epochs
following~\cite{Ma2019SelfSupervisedSS, Qu2020DepthCV}.
We train our method for 20 epochs and all others for 30.
This is to account for the increased training time using our method.
For Snapshot Ensemble,we follow the original paper \cite{Huang2017SnapshotET} 
and use the cyclic annealing scheduler from~\cite{Loshchilov2017SGDRSG}
with the same initial learning rate as before.  
we train for 5 epochs per cycle, and discard the worst snapshot, which leaves us with 5 snapshots.
All training is carried out on a single Tesla V100 GPU with the same batch size and random seed.
For data augmentation we apply a random horizontal flip with a probability of 0.5 
and a small color jitter of 0.02. 
\subsection{Metrics}\label{sec:eval_metrics}
\noindent 
\textbf{Depth prediction metrics.}
We evaluate depth completion performance using standard metrics \cite{Eigen2014DepthMP}.
Specifically, we report MAE, RMSE and accuracy ($\delta$-threshold) on depth.
Due to space limitation, we only report $\delta_1 < 1.25$.

\halfline
\noindent 
\textbf{Uncertainty estimation metrics.}
Unlike depth prediction which can be compared to ground truth, 
the true probability density function of depth is not available. 
This makes evaluating uncertainty estimates a difficult task in itself.
We describe three popular metrics from the literature 
for evaluating uncertainty estimates.
Note that each metric has its own advantages and drawbacks,
we seek to provide a more comprehensive evaluation by reporting all three.


\halfline
\noindent 
\textbf{1) Area Under the Sparsification Error curve (AUSE)$\downarrow$.} 
Sparsification plots \cite{Aodha2013LearningAC} are commonly used 
for measuring the quality of uncertainty estimates. 
Given an error metric (\eg MAE), 
we sort the prediction errors by their uncertainty in descending order 
and compute the error metric repeatedly by 
removing a fraction (\eg 1\%) of the most uncertain subset.
An oracle sparsification curve is obtained by sorting using the true prediction errors.
AUSE is the area between the sparsification curve and the oracle curve.
This normalizes the oracle out and can be used to compare different methods \cite{Ilg2018UncertaintyEA}.
Note that AUSE is a \textit{relative} measure of uncertainty quality, 
since its computation relies on the \textit{order} of predicted uncertainties.

\halfline
\noindent 
\textbf{2) Area Under the Calibration Error curve (AUCE)$\downarrow$.} 
For an \textit{absolute} measure of uncertainty estimation quality, 
\cite{Gustafsson2020EvaluatingSB} proposes to generalize the 
\textit{Expected Calibration Error (ECE)}~\cite{Guo2017OnCO} metric to regression.
For Laplace distributions, given mean $\mu$ and variance $\sigma^2$, 
we construct prediction intervals $\mu \pm \Psi^{-1}(\frac{p+1}{2})b$ for $p \in (0, 1)$, 
where $\Psi$ is the CDF of the unit Laplace distribution.
For each value of $p$, we compute the proportion of pixels $\hat{p}$ 
for which the true target falls within the predicted interval.
For a well-calibrated model, $\hat{p}$ should closely match $p$.
The \textit{Calibration Error curve} is defined as $|p - \hat{p}|$, 
and AUCE is the area under this curve.
Like ECE, AUCE is \textbf{not} a \textit{proper} scoring rule \cite{Ovadia2019CanYT},
as there exists trivial solutions which yield perfect scores. 

\halfline
\noindent 
\textbf{3) Negative Log-likelihood (NLL)$\downarrow$.}
NLL~\eqref{eq:loss_nll} is commonly used to evaluate 
the quality of model uncertainty on a held-out dataset \cite{Ovadia2019CanYT}. 
It is a \textit{proper} scoring rule \cite{Gneiting2007StrictlyPS}, 
but over-emphasizes tail probabilities \cite{Candela2005EvaluatingPU}
and cannot fully capture posterior in-between uncertainty \cite{Thakur2020LearnedU}.

\begin{table*}[t]
\centering
\footnotesize
\begin{tabular}{|c|c|c|ccc|ccc|ccc|ccc|}
\hline
\multicolumn{3}{|c|}{Trained with 5\%} 
& \multicolumn{6}{c|}{VKITTI2} 
& \multicolumn{6}{c|}{NYU-V2} \\ \hline 
Input & Method & \% 
& MAE & RMSE & $\delta_1$ & AUSE & AUCE & NLL 
& MAE & RMSE & $\delta_1$ & AUSE & AUCE & NLL
\\ \hline\hline

rgbd & snap~\cite{Huang2017SnapshotET} & 5\% 
& 1.192 & 3.267 & 95.59 & 0.445 & 0.170 & -0.714
& 0.061 & 0.126 & 99.35 & 0.036 & 0.202 & -1.390
\\ 

rgbd & snap+log~\cite{Kendall2017WhatUD} & 5\% 
& 1.271 & 3.432 & 95.33 & 0.142 & \textbf{0.117} & -1.582
& 0.058 & 0.123 & 99.32 & 0.018 & 0.256 & -1.596
\\

rgbd & log~\cite{Nix1994EstimatingTM} & 5\%  
& 1.318 & 3.423  & 95.37 & 0.149 & 0.125 & -1.421
& 0.057 & 0.121  & 99.34 & 0.018 & 0.210  & -1.783
\\

rgbd & bdbf & 5\%  
& \textbf{0.703} & \textbf{2.925} & \textbf{97.88} & \textbf{0.110} & 0.136 & \textbf{-2.596} 
& \textbf{0.026} & \textbf{0.082} & \textbf{99.64} & \textbf{0.007} & \textbf{0.039} & \textbf{-3.151} 
\\ \hline 
\end{tabular}
\caption{
Quantitative results of all methods trained and tested with 5\% sparsity on VKITTI2 and NYU-V2. 
}
\label{tab:mid_all_5p}
\end{table*}

\subsection{Results}

\begin{table*}[t]
\centering
\footnotesize
\begin{tabular}{|c|c|c|ccc|ccc|ccc|ccc|}
\hline
\multicolumn{3}{|c|}{Trained with 500} 
& \multicolumn{6}{c|}{VKITTI2} 
& \multicolumn{6}{c|}{NYU-V2} \\ \hline
Input & Method & \# 
& MAE & RMSE & $\delta_1$ & AUSE & AUCE & NLL 
& MAE & RMSE & $\delta_1$ & AUSE & AUCE & NLL 
\\ \hline\hline

rgbd & snap & 500 
& 2.312 & 5.403 & 90.14 & 0.459 & 0.229  & -0.207
& 0.096 & 0.206 & 97.53 & 0.053 & 0.261 & 0.211 
\\ 

rgbd & snap+log & 500 
& 2.396 & 5.571 & 89.88 & \textbf{0.273} & 0.036 & -1.150
& 0.095 & 0.213 & 97.44 & 0.025 & 0.205 & -1.393
\\

rgbd & log & 500 
& 2.492 & 5.800 & 89.13 & 0.299 & 0.095 & -0.906
& 0.097 & 0.212 & 97.44 & 0.025 & 0.152 & -1.502
\\ 

rgbd & bdbf & 500 
& \textbf{2.015} & \textbf{4.994} & \textbf{92.71} & 0.392 & \textbf{0.014} & \textbf{-1.215}
& \textbf{0.064} & \textbf{0.166} & 98.46 & \textbf{0.021} & 0.030 & \textbf{-2.199} 
\\ 

\hline
rgb  & bdbf & 500 
& 2.569 & 5.642 & 88.67 & 0.481 & 0.015 & -0.979
& 0.098 & 0.199 & \textbf{98.48} & 0.030 & \textbf{0.014} & -1.689
\\ \hline \hline

rgb$^\dag$  & log & 0 
& 6.758 & 11.78 & 61.48 & 1.591 & 0.291 & 2.407
& \textbf{0.366} & \textbf{0.561} & \textbf{75.08} 
& \textbf{0.161} & 0.191 & \textbf{0.187}
\\

rgb & bdbf & 0 
& \textbf{5.809} & \textbf{9.610} & \textbf{62.78} 
& \textbf{1.381} & \textbf{0.264} & \textbf{0.809} 
& 0.664 & 0.944 & 47.76
& 0.245 & \textbf{0.044} & 0.459
\\ \hline
\end{tabular}
\caption{
Quantitative results of all methods trained and tested with 500 sparse depths
and our proposed method tested with no sparse depths 
compared to a monocular depth prediction baseline($\dag$).
\textit{rgbd} under the input column indicates the basis network uses the sparse depths 
scaffolding approach from~\cite{Wong2020UnsupervisedDC}, 
whereas \textit{rgb} uses color image as basis network input only.
}
\label{tab:low_all_500_0}
\end{table*}

\begin{figure}
\centering
\includegraphics[width=\linewidth]{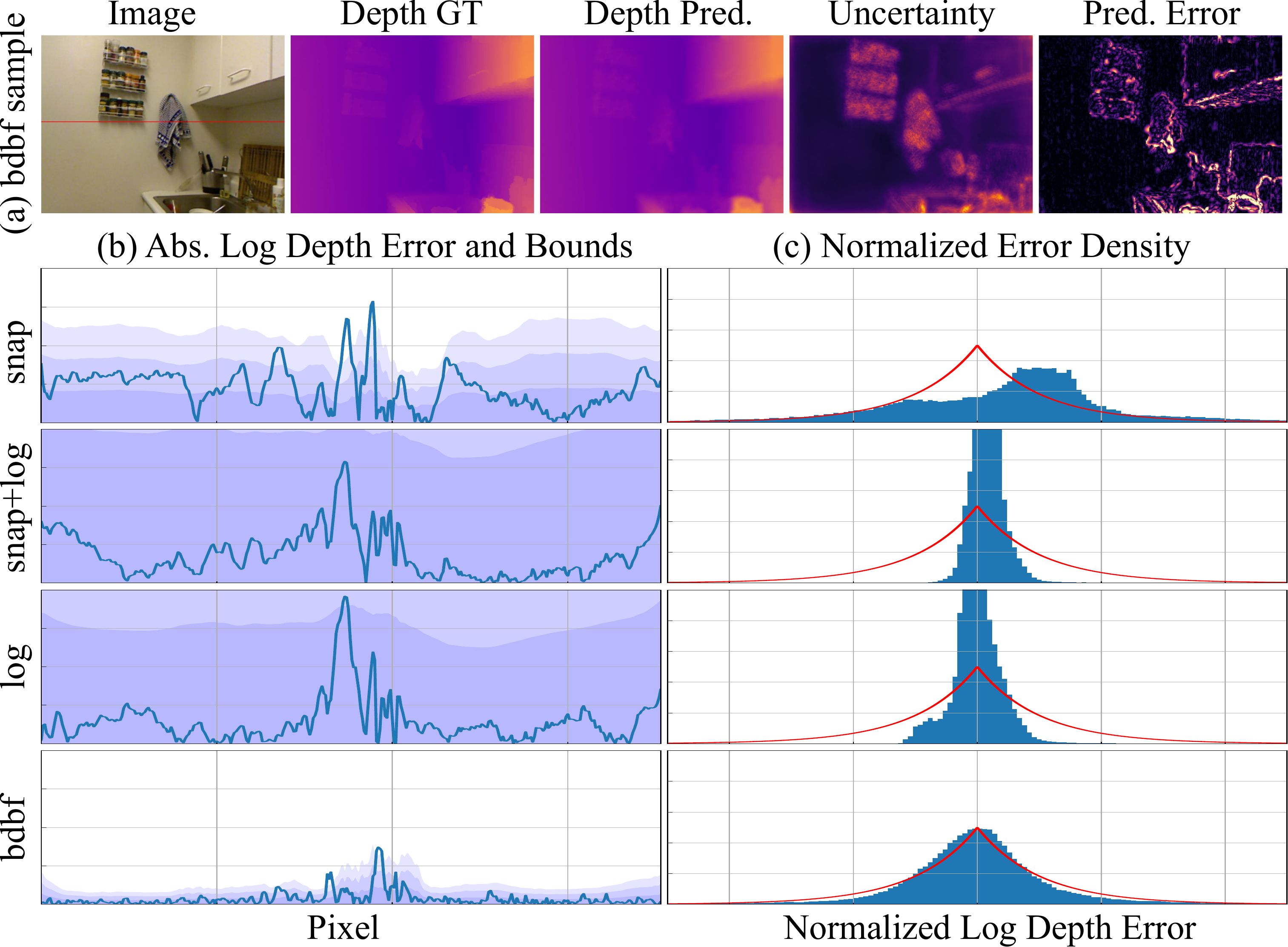}
\caption{
(a) Qualitative results of \textit{bdbf} on one test image from NYU-V2.
(b) Absolute log depth error (blue line) and $3b$ bounds (blue shades) 
for a single row of pixels (red line) from the image.
(c) Normalized error density of the entire image compared to a unit Laplace distribution
(red line).
All axes are of the same scale within each column.
}
\label{fig:mid_nyu_bound_density}
\end{figure}

\begin{figure}[t]
\centering
\includegraphics[width=\linewidth]{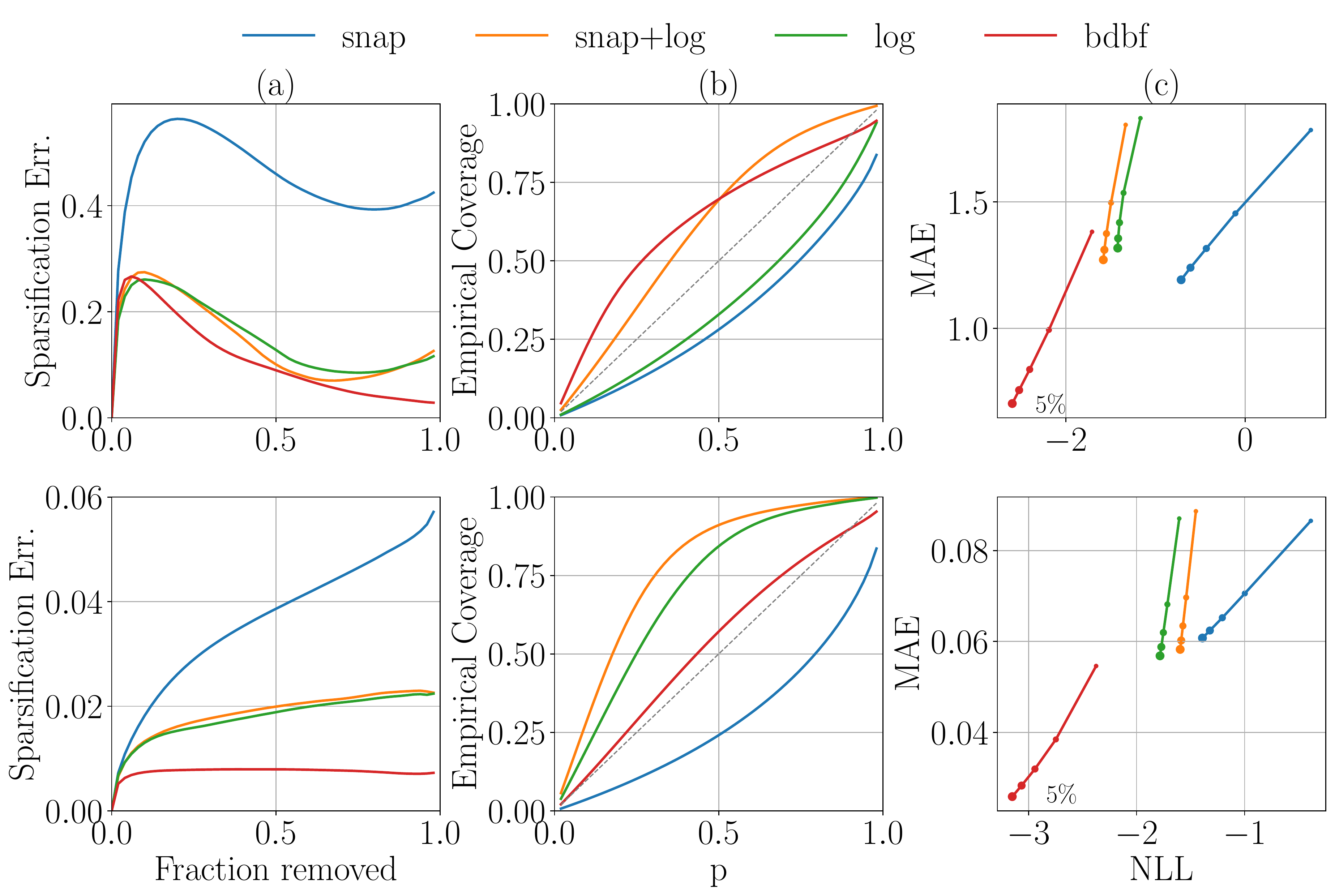}
\caption{
(a) Sparsification error, (b) calibration error
and (c) sparsity change plots of NLL \vs MAE 
with 5\% sparsity on VKITTI2 (top) and NYU-V2 (bottom). 
Sparsification and calibration plots are generated using 5\% test sparsity.
Sparsity change plots are generated with varying test sparsity from 5\% to 1\%.
}
\label{fig:mid_ause_auce_2d}
\end{figure}

\halfline 
\noindent
\textbf{Mid-density Depth Completion.}
In this setting, we train all methods with 5\% sparsity.
Table~\ref{tab:mid_all_5p} shows quantitative results 
when testing on the same dataset under the same sparsity level.
This is considered an \textit{in-distribution} test.
We see significant improvements of our method in almost all metrics.
Figure~\ref{fig:mid_vk2_qual} shows qualitative results of 
one sample from the VKITTI2 test set.
Compared to others, \textit{bdbf} not only predicts higher quality depths
but also sharper uncertainties that closely match the true prediction errors.
This indicates that the learned depth bases in ours 
is expressive for predicting both depth and uncertainty.

We take a closer look at one sample from the NYU-V2 test set 
in Figure~\ref{fig:mid_nyu_bound_density} (a), 
by plotting the absolute prediction error in log space $e = |\mu - z|$ 
and uncertainty bound $b = \sigma / \sqrt{2}$ 
for one row of pixels in Figure~\ref{fig:mid_nyu_bound_density} (b),
and the normalized error density $\widetilde{e} = e / b$ 
on the entire image in Figure~\ref{fig:mid_nyu_bound_density} (c).
\textit{bdbf} is the only method where the bound traces the general shape of the prediction error 
and whose normalized error density resembles that of a unit Laplace distribution. 
\textit{snap} fails to capture the underlying error distribution.
\textit{log} and \textit{snap+log} produce decent relative uncertainties (AUSE) 
but are not well-calibrated (AUCE).
Figure~\ref{fig:mid_ause_auce_2d} (a) (b) show sparsification error 
and calibration plots for the in-distribution test on both datasets, 
which are used to compute AUSE and AUCE respectively.
We see that predictive method (\textit{log}) performs similar 
to its ensemble variant \textit{snap+log} 
and both are better than the pure ensemble method, \textit{snap}.
This is consistent with findings in \cite{Ilg2018UncertaintyEA, Poggi2020OnTU}.

We also evaluate all methods under the effect of
distributional (dataset) shift~\cite{Ovadia2019CanYT}.
Here we mainly focus on the following two aspects: 
\textbf{sparsity change} and \textbf{domain shift}. 
For sparsity change within mid-density, we take models trained on 5\% sparsity 
and test on varying sparsity level from 5\% to 1\%.
Results are shown in Figure~\ref{fig:mid_ause_auce_2d} (c).
Note that these plots reflect how each method performs on two axes 
in terms of uncertainty estimation (NLL) and depth prediction (MAE), 
where better methods should reside closer to the lower left corner.
We see that performance of all methods degrade 
in a similar manner with deceasing sparsity, 
which is largely due to the sparse depth scaffolding approach we choose.
However, \textit{bdbf} stands out with its 1\% result 
better than the 5\% results of its competitors.
We refer the reader to our supplementary material for results on domain shift.


\begin{figure}
\centering
\includegraphics[width=\linewidth]{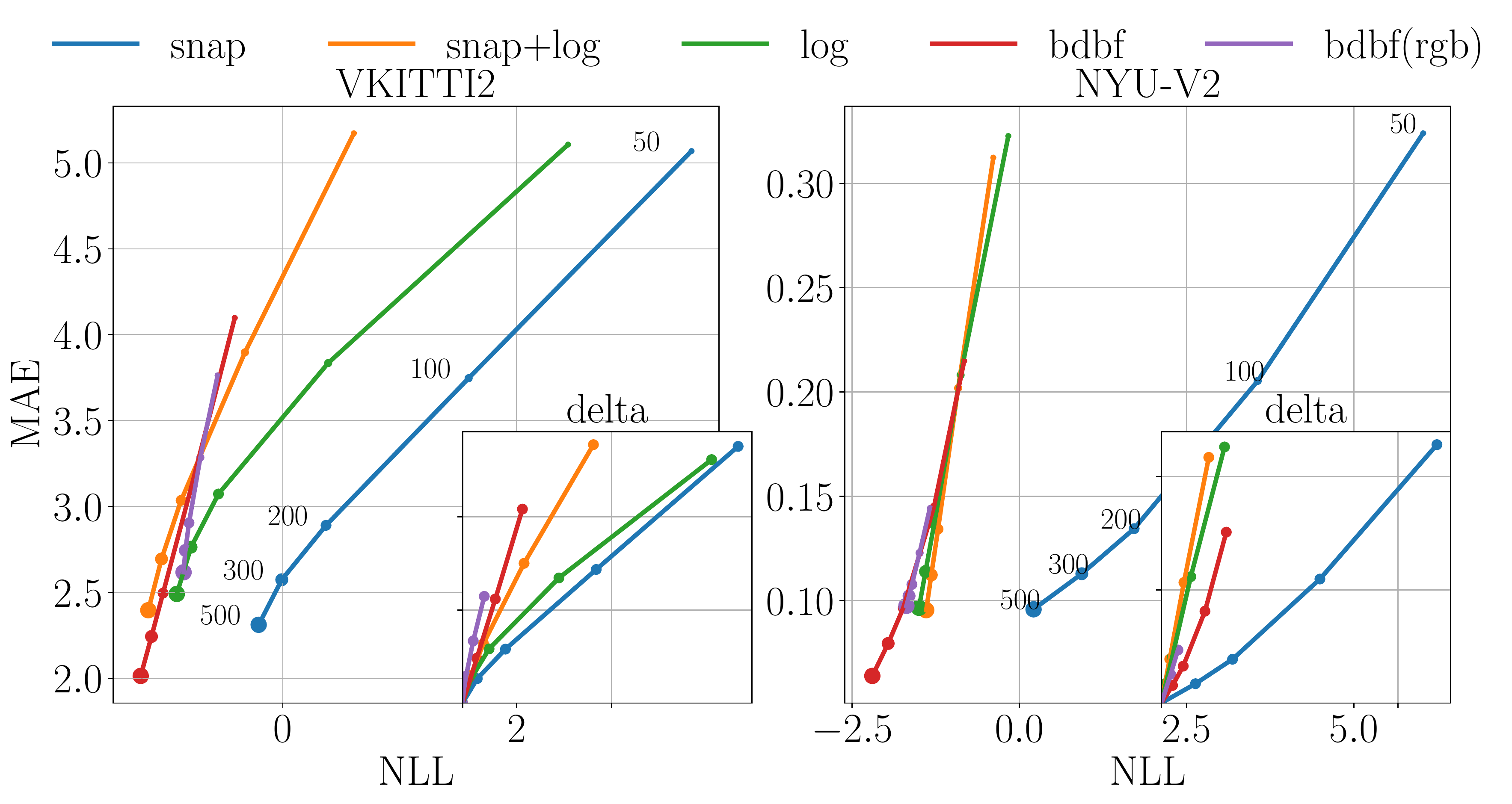}
\caption{
Sparsity change plots of all methods trained with 500 sparse depths 
and test with various sparsity from 500 to 50.
The small subplots show how the performance of each method changes 
with decreasing sparsity \wrt 
the performance of their in-distribution test (500).
Shorter lines indicate better sparsity-invariance.
}\label{fig:low_nll_mae_2d}
\end{figure}

\halfline
\noindent
\textbf{Low-density Depth Completion.}
In this setting, we train all methods with 500 sparse points, 
which is roughly 0.5\% sparsity given our image size.
We also introduce a slight variation of our method \textit{bdbf(rgb)}
which only uses sparse depths at the fitting stage (not as network input).
Because at very low sparsity levels (\eg 50 points), 
the scaffolding method we use for depth interpolation \cite{Wong2020UnsupervisedDC}
struggles to recover the scene structure which impacts the 
performance of all \textit{rgbd} methods.

The top half of Table \ref{tab:low_all_500_0} shows 
the in-distribution test of all methods.
Among the four \textit{rgbd} methods,
\textit{bdbf} again outperforms the rest by a large margin.
\textit{bdbf(rgb)}, despite not utilizing the rich information 
provided by the interpolated depths,
performs on-par with the baselines.
The real advantage of this approach is that it does not suffer from 
the artifacts caused by poor depth interpolation in the very low sparsity regime,
which makes it sparsity-invariant.
This claim is verified in Figure \ref{fig:low_nll_mae_2d}, 
which shows how each method's performance deteriorates with decreasing sparsity.
It is shown in the small subplots that \textit{bdbf(rgb)} 
is able to maintain good performance even with as few as 50 points.

\begin{figure}
\centering
\includegraphics[width=0.9\linewidth]{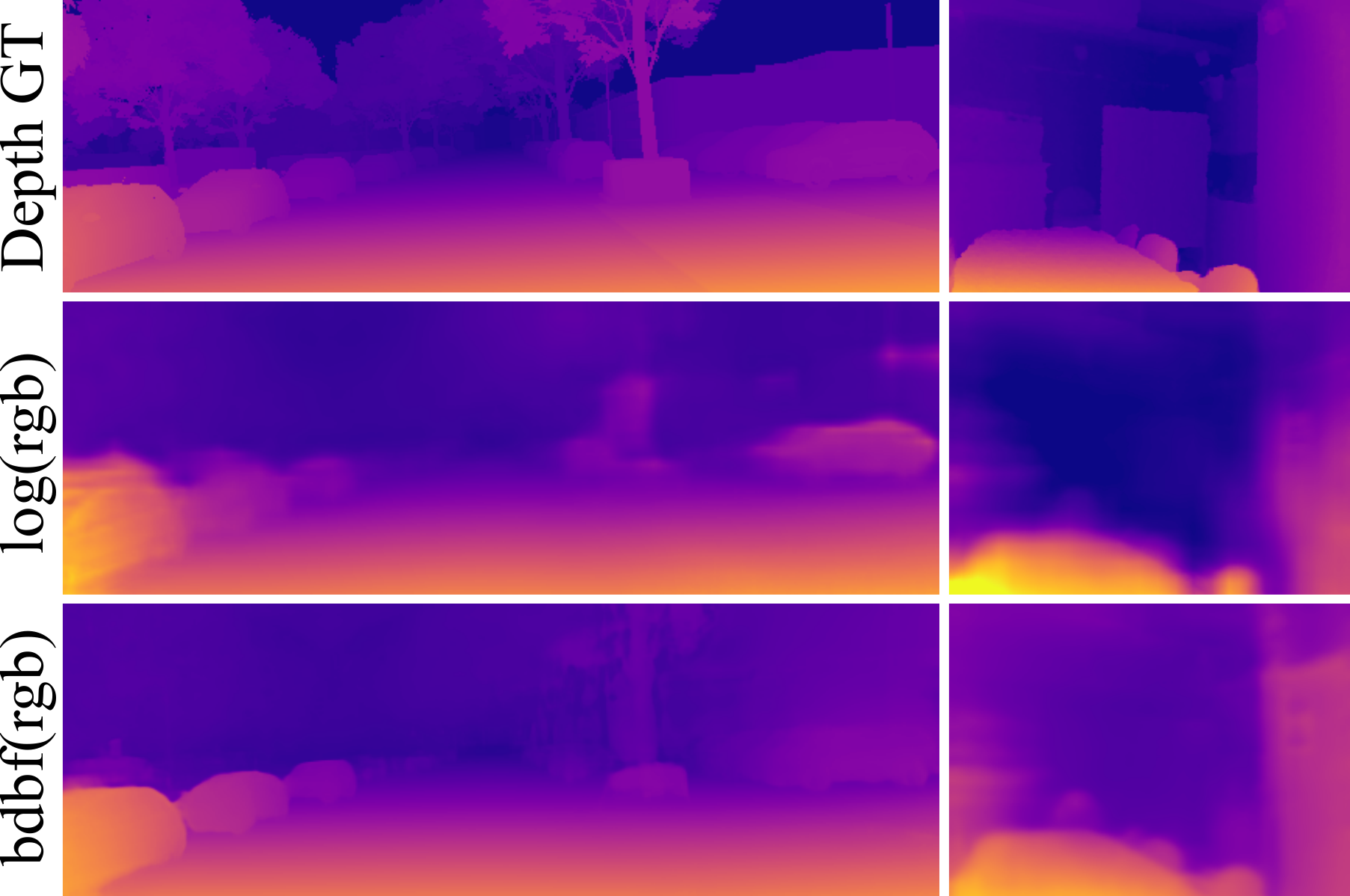}
\caption{
Qualitative results of our method tested with 0 sparse points.
\textit{log(rgb)} is trained as a monocular depth prediction network with NLL loss, 
which serves as a baseline.
\textit{bdbf(rgb)} is trained with 500 sparse depths.
}
\label{fig:low_nyu_vk2_n0}
\end{figure}

Finally, we test \textit{bdbf(rgb)} with no sparse depths,
which relies only on the shared prior to make a prediction.
We ignore all \textit{rgbd} methods because with nothing 
to interpolate the network outputs poor solutions.
We thus only compare to another baseline \textit{log(rgb)},
which is trained for monocular depth prediction with NLL loss.
Note that \textit{bdbf(rgb)} and \textit{log(rgb)} have 
exactly the same architecture (except for the last layer) 
and number of parameters.
We see that \textit{bdbf(rgb)} 
produces sharper depth than the baseline 
as shown in Figure~\ref{fig:low_nyu_vk2_n0}.
Quantitative results can be found in 
the last two rows of Table~\ref{tab:low_all_500_0}.
The difference in performance of our method between two datasets 
is due to the distribution of the data: 
VKITTI2 contains mainly sequential driving videos,
which gives a sharply peaked prior;
whereas data from NYU-V2 are taken from a wide variety of scenes 
with different viewing angles, hence a less informative one.
These results show that our learned depth bases and shared prior
contain geometric information about the scene conditioned on the image
and can be used under extreme conditions without catastrophic failure. 

\begin{table}[t]
\centering
\footnotesize
\begin{tabular}{|c | c | c | c | c | c|}
\hline
Method & iRMSE & iMAE & RMSE & MAE \\ \hline\hline
S2D~\cite{Ma2019SelfSupervisedSS} & 2.80 & 1.21 & 814.73 & 249.95 \\
Gansbeke~\cite{Gansbeke2019SparseAN} & 2.19 & 0.93 & 772.87 & 215.02 \\
DepthNormal\cite{Xu2019DepthCF} & 2.42 & 1.13 & 777.05 & 235.17 \\
DeepLiDAR \cite{Qiu2018DeepLiDARDS} & 2.56 & 1.15 & 758.38 & 226.50 \\
FuseNet \cite{Chen2019LearningJ2} & 2.34 & 1.14 & 752.88 & 221.19 \\
CSPN++~\cite{Cheng2020CSPNLC} & 2.07 & 0.90 & 743.69 & 209.28 \\ 
NLSPN~\cite{Park2020NonLocalSP} & 1.99 & 0.84 & 741.68 & 199.59  \\
GuideNet~\cite{Tang2019LearningGC}  & 2.25 & 0.99 & 736.24 & 218.83 \\ \hline
bdbf (ours) & 2.37 & 0.89 & 900.38 & 216.44 \\ \hline
\end{tabular}
\caption{
Comparison with selected methods on the official KITTI depth completion test set.
}
\label{tab:kitti_sota}
\end{table}

\begin{table}[t]
\centering
\footnotesize
\begin{tabular}{|c | c| c | c | c | c |} \hline
Method & MAE & RMSE & AUSE & AUCE & NLL \\ \hline \hline
NCNN-L2~\cite{Eldesokey2018PropagatingCT} & 258.68 & 954.34 & 0.70 & - & - \\
pNCNN~\cite{Eldesokey2020UncertaintyAwareCF} & 283.41 &  1237.65 & 0.055 & - & - \\
pNCNN-Exp & 251.77 & 960.05 & 0.065 & - & - \\ \hline
bdbf (ours) & 206.70 & 876.76 & 0.057 & 0.23 & -2.68 \\
\hline
\end{tabular}
\caption{
Comparison with variations of pNCNN~\cite{Eldesokey2020UncertaintyAwareCF} on accuracy and uncertainty on the official KITTI validation set (with groundtruth). 
Note that pNCNN is \textit{unguided}.
}
\label{tab:pncnn}
\end{table}

\noindent
\textbf{Further Comparisons.}
While our focus is on evaluating the quality of our uncertainty estimation scheme, 
we also evaluate depth completion performance for completeness.
We trained our method with a ResNet34 encoder~\cite{He2016DeepRL} and 
applied it to the KITTI depth completion benchmark
with results shown in Table \ref{tab:kitti_sota}.
We compare our relatively simple Bayesian filtering scheme to SOTA methods that utilize either iterative refinement~\cite{Cheng2020CSPNLC, Park2020NonLocalSP}
or sub-networks with extra constraints~\cite{Xu2019DepthCF, Qiu2018DeepLiDARDS}.
Our method compares favorably on all measures except RMSE and we observe that this difference is due to a small number of mis-attributed pixels near depth discontinuities and the use of L1 loss only.
This suggests that these methods could be further improved by predicting 
initial depth and uncertainty estimates with our module.

We also compare with  pNCNN~\cite{Eldesokey2020UncertaintyAwareCF} 
as it is the only work that provides a quantitative evaluation of predicted uncertainties for depth completion.
Unfortunately, they only evaluate using a single metric, AUSE,
which we argue cannot completely capture the true quality of the uncertainty estimate.
Results are shown in Table \ref{tab:pncnn}, note that pNCNN is unguided 
and evaluation is done on the KITTI validation set, 
as groundtruth is required to compute uncertainty metrics.

\section{Conclusions}
In this paper, we extend Deep Basis Fitting for depth completion
under a principled Bayesian framework 
that outputs uncertainty estimates alongside depth prediction.
Compared to commonly used uncertainty estimation techniques,
our integrated approach is able to produce better uncertainty estimates 
while being data- and compute-efficient.
The benefit of being Bayesian is also demonstrated by 
the ability to handle very low-density sparse depths, 
a situation where the original DBF method struggles.
Our work allows a depth completion network
to be further integrated into robotics systems,
where Bayesian sensor fusion is the dominant approach.
\setcounter{section}{5}
\section{Network Architecture}

The specific network architecture is of little importance to our approach, 
since all methods use exactly the same basis network, initialization and random seeds.
Nevertheless, we list detailed architecture here for completeness.

We use the scaffolding approach detailed in \cite{Wong2020UnsupervisedDC} 
to interpolate sparse depths input. 
We then adopt an early fusion strategy
where the interpolated depth map 
goes through a single convolution block (with normalization and nonlinear activation) 
with stride 2
to be merged with the first stage feature map of the image encoder.
The fused feature map is then used as input to the subsequent stages of the encoder.
The encoder is an ImageNet-pretrained MobileNet-v2 \cite{Sandler2018MobileNetV2IR}
which output a set of multi-scale feature maps $\{E_i\}$ at each stage
with channels (16, 24, 32, 96, 320) in decreasing resolution.
These feature maps except for the last one are then used as skip connection to the decoder,
which eventually output another set of multi-scale features $\{D_i\}$ 
with channels (256, 192, 128, 64, 32) in increasing resolution.
We keep the encoder essentially intact, while using ELU \cite{Clevert2016FastAA} 
activation throughout the decoder.
This is due to the suggestion from \cite{Snoek2015ScalableBO}, 
where they found a smooth activation function produces better quality uncertainty estimates.
The decoder output $\{D_i\}$ are then used to generate the final multi-scale bases $\{\Phi_i\}$ 
with channels (2, 4, 8, 16, 32) in increase resolution.
They are then upsampled (via bilinear interpolation) to the input image resolution 
and concatenated together \cite{Qu2020DepthCV}.
This final 63-dimensional basis $\Phi$ (with a bias channel) is fed into 
either our proposed \textit{bdbf} module,
or a normal convolution to generate the latent prediction
before the final activation function $g$.

\section{Derivations}

In this section, we provide derivation of equations given in the main paper,
which roughly follows the same order of their original appearance.

\subsection{Marginal Likelihood}
The derivation of the full marginal likelihood function 
largely follows that from Chapter 3.5 in \cite{Bishop2006PatternRA}.

We start by writing the evidence function in the form 
\begin{align}
\begin{split}
&p(\mathbf{z} | \alpha, \beta)  \\
=& \int p(\mathbf{z} | \mathbf{w}, \beta) p(\mathbf{w}|\alpha) \mathrm{d}\mathbf{w} \\
=& |\bm{\Sigma}_0|^{-1/2}
\left( \frac{\beta}{2\pi} \right)^{\frac{N}{2}}
\left( \frac{\alpha}{2\pi} \right)^{\frac{M}{2}}
\int \exp \left\{-\frac{1}{2}E(\mathbf{w}) \right\} \mathrm{d} \mathbf{w}
\end{split}
\end{align}
where 
\begin{align}
E(\mathbf{w}) = 
\beta  \| \mathbf{z} - \bm{\Phi}\mathbf{w} \|^2 +
\alpha \| \mathbf{w} - \mathbf{m}_0\|^2_{\mathbf{\bm{\Sigma}_0}}
\end{align}
Completing the square over $\mathbf{w}$ we have
\begin{align}
E(\mathbf{w}) = E(\mathbf{m}) + \|\mathbf{w} - \mathbf{m}\|^2_{\bm{\Sigma}}
\end{align}
The integral term is evaluated by 
\begin{align}
\begin{split}
& \int \exp \left\{ -\frac{1}{2} E(\mathbf{w}) \right\} \mathrm{d}\mathbf{w}\\
=& \exp\left\{-\frac{1}{2}E(\mathbf{m}) \right\} (2\pi)^{M/2} |\bm{\Sigma}| ^{-1/2}
\end{split}
\end{align}
We can then write the log marginal likelihood as
\begin{align}\label{eq:supp_marg_like}
\begin{split}
\ln p(\mathbf{z}| \alpha, \beta) 
&= \frac{1}{2} (
N \ln \beta + M \ln \alpha - N \ln(2 \pi)  \\ 
& \qquad - E(\mathbf{m})  + \ln |\bm{\Sigma}| - \ln |\bm{\Sigma}_0|)
\end{split}
\end{align}

\subsection{Normalized Estimation Error Squared (NEES)}

NEES was originally used to indicate the performance of a filter~\cite{BarShalom2001EstimationWA},
\eg in a tracking application.
It is defined as
\begin{align}
\varepsilon_k
= (\mathbf{x}_k - \hat{\mathbf{x}}_{k|k})^\top \mathbf{P}^{-1}_{k|k}
(\mathbf{x}_k - \hat{\mathbf{x}}_{k|k})
\end{align}
where $\mathbf{x}$ is the true state vector, 
$\mathbf{P}$ is the state covariance matrix,
and $(\hat{\cdot})_{k|k}$ denotes the estimated posterior at time $k$.
It is also used in Simultaneous Localization and Mapping algorithms (SLAM)
to measure filter consistency~\cite{Bailey2006ConsistencyOT}. 
Specifically, the average NEES over N Monte Carlo runs is used.
Under the assumption that the model is correct (approximately linear Gaussian),
$\varepsilon_k$ is $\chi^2$ distributed with $\dim(\mathbf{x}_k)$.
Then the expected value of $\varepsilon_k$ should be
\begin{align}
E[\varepsilon_k] = \dim(\mathbf{x}_k)
\end{align}
For depth estimation, 
the state is the predicted depth at each pixel, 
which is one-dimensional.
Therefore, we expect a consistent depth estimator to have an average NEES of 1.

NEES can also be extended to a Laplace distribution
which is used in this work due to the choice of L1 loss. 
Given $z \sim \mathrm{Laplace}(\mu, b)$ 
and the fact that $\frac{2}{b}|z -\mu| \sim \chi^2(2)$,
the expected NEES (with one degree of freedom) is
\begin{align}
E[\varepsilon] = E\left[\left(\frac{z - \mu}{b}\right)^2 \right] = 1
\end{align}

\subsection{Re-estimation Equations}

Here we provide derivation of the re-estimation equations
used in the EM step during inference~\cite{Bishop2006PatternRA}.
In the E step, we compute the posterior distribution of $\mathbf{w}$ 
given the current estimation of the parameters $\alpha$ and $\beta$.
In the M step, we maximize the expected complete-data log likelihood 
with respect to $\alpha$ and $\beta$ and re-iterate until convergence.
The convergence criteria is met when the change in relative magnitude of $\beta$ is smaller than 1\%.

The complete-data log likelihood function is given by
\begin{align}
\ln p(\mathbf{z}, \mathbf{w} | \alpha, \beta)
= \ln p(\mathbf{z} | \mathbf{w}, \beta) 
+ \ln p(\mathbf{w} | \alpha)
\end{align}
with
\begin{align}
p(\mathbf{z} | \mathbf{w}, \beta)
&= \mathcal{N}(\mathbf{z} | \bm{\Phi}\mathbf{w}, \beta^{-1} \mathbf{I})  \\
p(\mathbf{w} | \alpha) 
&= \mathcal{N}(\mathbf{w} | \mathbf{m}_0, \alpha^{-1} \bm{\Sigma}_0)
\end{align}
The expectation \wrt the posterior distribution of $\mathbf{w}$ is 
\begin{align}
\begin{split}
&E[\ln p(\mathbf{z}, \mathbf{w} | \alpha, \beta)] \\
=& \frac{M}{2} \ln\left( \frac{\alpha}{2\pi} \right) - \frac{1}{2} \ln |\bm{\Sigma}_0| 
- \frac{\alpha}{2} \mathbb{E} [\| \mathbf{w} - \mathbf{m}_0 \|^2_{\bm{\Sigma}_0}]
\\
+& \frac{N}{2}\ln \left(\frac{\beta}{2\pi} \right) 
- \frac{\beta}{2} \mathbb{E} [ \|  \mathbf{z} - \bm{\Phi} \mathbf{w} \|^2]
\end{split}
\end{align}
Setting the derivatives with respect to $\alpha$ and $\beta$ to zero
gives us the re-estimation equations.
\begin{align}
\begin{split}
\alpha
&= \frac{M}{\mathbb{E} [ \| \mathbf{w}- \mathbf{m}_0 \|^2_{\bm{\Sigma}_0}]} \\
&= \frac{M}{\tr(\bm{\Sigma}_0^{-1}\bm{\Sigma}) + \| \mathbf{m}- \mathbf{m}_0 \|^2_{\bm{\Sigma}_0}}
\end{split} \\
\begin{split}
\beta 
&= \frac{N}{\mathbb{E} [ \| \mathbf{z} - \bm{\Phi} \mathbf{w} \|^2]} \\
&= \frac{N}{ \tr(\bm{\Phi}^\top \bm{\Phi} \bm{\Sigma}) + \|\mathbf{z} - \bm{\Phi}\mathbf{m} \|^2}
\end{split}
\end{align}

\section{More Results}
\begin{figure*}[t]
\centering
\includegraphics[width=\linewidth]{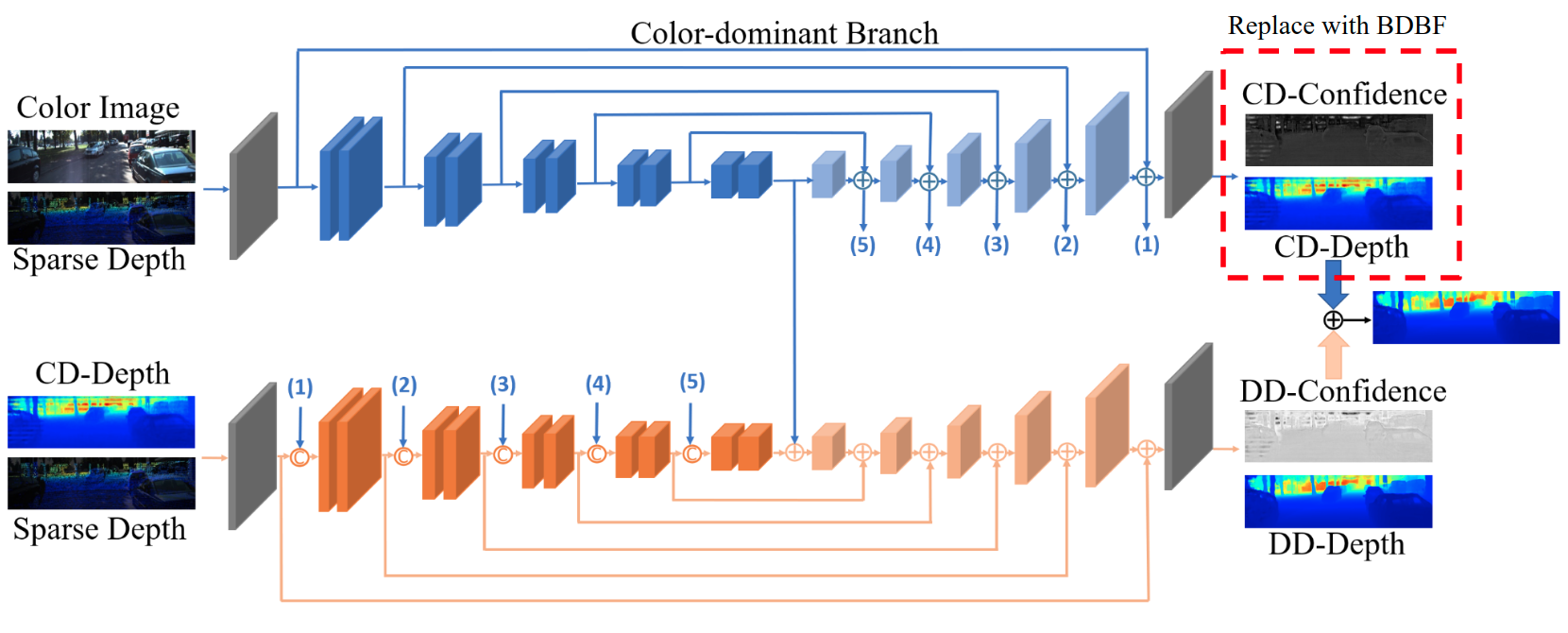}
\caption{The backbone of PENet~\cite{Hu2021PENetTP}, which is called ENet, 
and our modification. Figure taken directly from the paper.}
\label{fig:penet_backbone}
\end{figure*}

\begin{figure}[t]
\centering
\includegraphics[width=\linewidth]{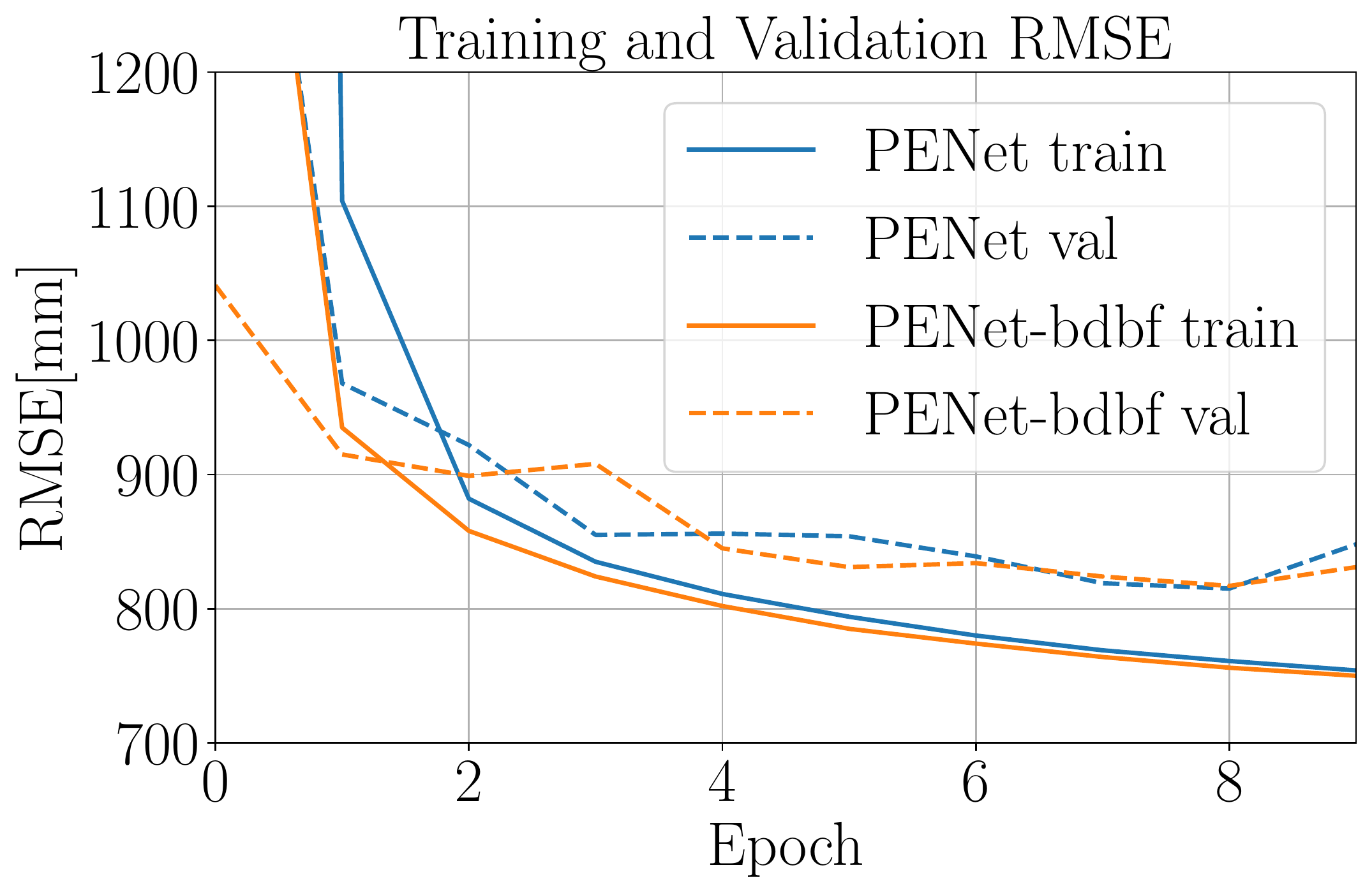}
\caption{Training and validation RMSE of PENet and PENet-bdbf 
after training for 10 epochs on KITTI Depth Completion dataset.}
\label{fig:penet_result}
\end{figure}

\subsection{Using BDBF as a General Component}
Here we verify our claim that \textit{bdbf} can be used as a general, drop-in, component 
in a depth completion network.
The best published method on the KITTI depth completion benchmark 
at the time of this writing is PENet~\cite{Hu2021PENetTP}.
It has a fully-convolutional backbone called ENet that 
also ranks third on the associated leaderboard. 
For details of their approach, we refer the reader to the original paper~\cite{Hu2021PENetTP}.
We only discuss the minimal changes that we made to add \textit{bdbf}.

Figure~\ref{fig:penet_backbone} shows the network architecture for ENet.
The color-dominant (CD) branch uses a convolutional layer to predict CD-depth and CD-confidence.
We simply replaced this convolutional layer with our \textit{bdbf} module.
However, we needed to normalize our variance prediction to conform to their notion of confidence.
This was done by inverting the standard deviation 
and passing it through a \texttt{Tanh} nonlinearity to constrain the result to lie between 0 and 1.
We then trained both versions for 10 epochs.
Training and validation results are show in Figure~\ref{fig:penet_result}.
We see that by using \textit{bdbf}, we achieve slightly better validation RMSE 
and faster convergence.
\subsection{Inference Time}

In table \ref{tab:mid_runtime}, we show inference time of all methods
when running on an image of size $320\times 240$ with 5\% sparsity levels.
All ensemble methods (\textit{snap} / \textit{snap+log}) use 5 snapshots
as described in the main paper.
They take roughly 5$\times$ times longer for an inference pass 
compared to \textit{log}, which is the fastest among all.
For \textit{bdbf} we use a maximum iteration of 8, 
but we observe convergence within 2 iterations for this particular density.
\textit{bdbf} is slower than \textit{log} due to the need to 
solve a small linear system and the iterations required for EM,
but it is has slower latency and smaller memory footprints 
compared to ensemble methods.
Note that during inference, we move expensive computations in our method
like Cholesky and LU decomposition from GPU to CPU and then move the results back.
Since this is only done in evaluation mode, 
no back-propagation is needed and thus
no computation graph broken.
The time needed to transfer small tensors ($\dim(\mathbf{w})\times \dim(\mathbf{w})$)  
between host and device memory is negligible 
compared to carrying out those operations on GPU.
Without these changes, \textit{bdbf} runs as slow as \textit{snap}.

\begin{table}[h]
\centering\footnotesize
\begin{tabular}{|c|c|c|c|c|c|}
\hline
Inference & snap  & snap+log & log   & dbf   & bdbf \\ \hline
Time {[}ms{]} & 68.84 & 66.17 & 16.68 & 21.84 & 23.97  \\ \hline
\end{tabular}
\caption{
Inference time of all methods with 5\% sparsity. Image resolution is $320\times 240$.
}
\label{tab:mid_runtime}
\end{table}

\subsection{Estimator Consistency}

\begin{figure}
\centering
\includegraphics[width=\linewidth]{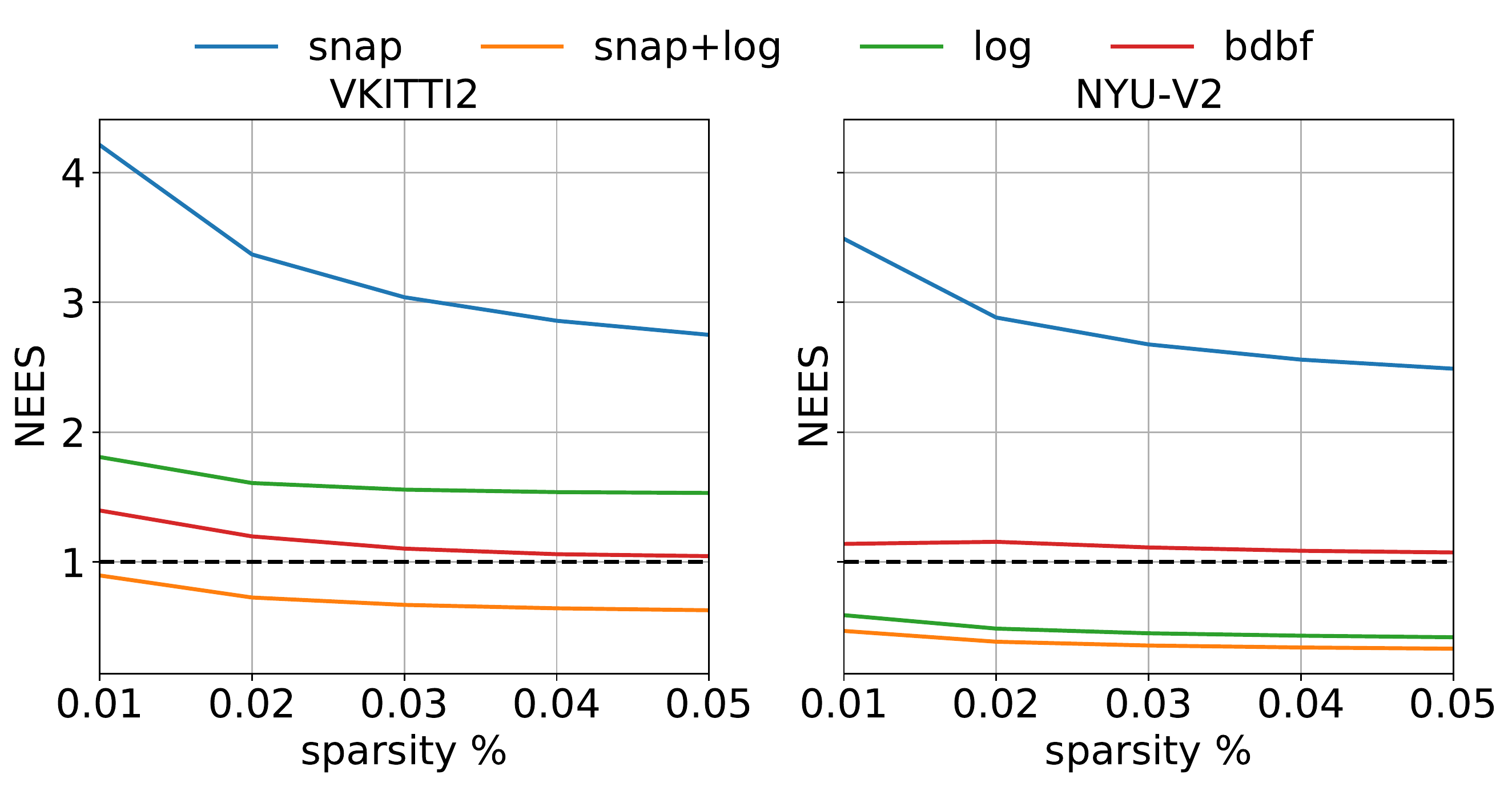}
\caption{
NEES scores for all methods with 5\% sparsity.
Closer to 1 (dashed black line) means more consistent.
}
\label{fig:supp/mid_nees}
\end{figure}

\begin{figure}
\centering
\includegraphics[width=\linewidth]{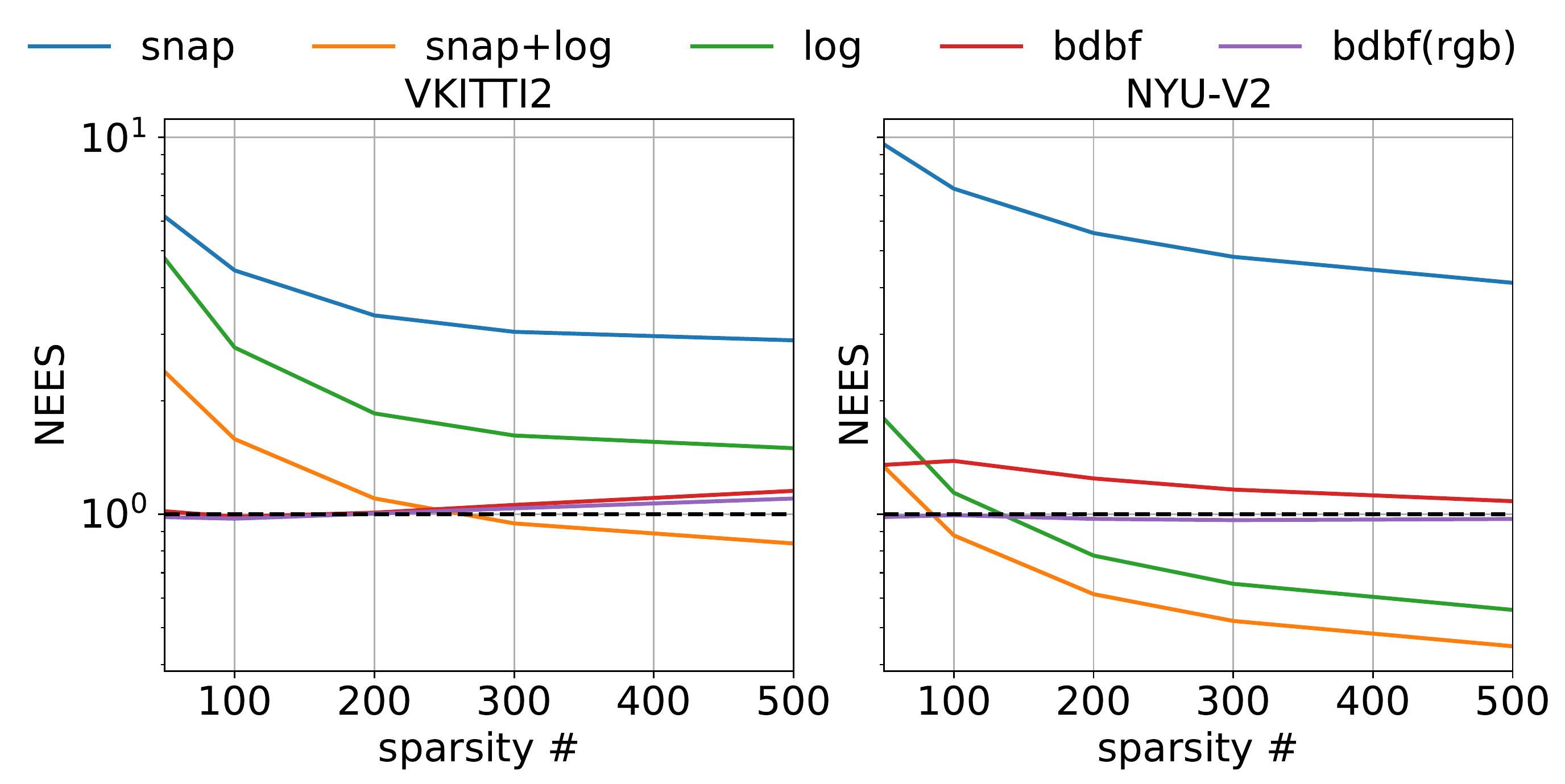}
\caption{
NEES scores for all methods with 500 sparse depths.
Closer to 1 (dashed black line) means more consistent.
Note the log scale on y-axis.
}
\label{fig:supp/low_nees}
\end{figure}

Here we report average NEES scores for all methods on various datasets.
NEES is not used as a metric in our evaluation 
since our method explicitly uses it for uncertainty calibration
and would render the comparison unfair.
We present it here only to show the efficacy of our uncertainty calibration scheme.
Figure \ref{fig:supp/mid_nees} and \ref{fig:supp/low_nees} show NEES of all methods
under mid- and low- density with varying sparsity level.
We see that our methods are the most consistent of all 
and remain relatively consistent even with sparsity change.
Note that this consistency-based calibration can not be applied to other methods, 
because we did not observe the same amount of over- or under-confidence from them.
While calibration of other baseline models is feasible, 
doing so would require extra data and thus not considered in this work.

\subsection{Shared Prior}

\begin{figure}[h]
\centering
\includegraphics[width=\linewidth]{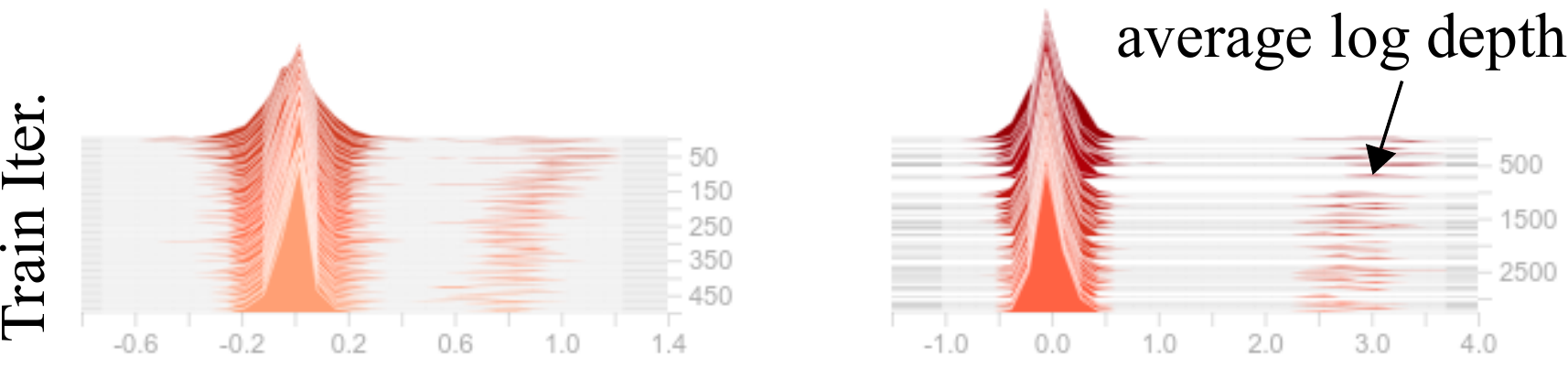}
\caption{
Tensorboard records of Weight histograms of two training sessions 
on NYU-V2 (left) and VKITTI2 (right).
}
\label{fig:supp/weight_hist}
\end{figure}

Figure \ref{fig:supp/weight_hist} supports our claim in the main paper
about the assumption of a shared prior across samples.
Here we show the weight histograms of two training sessions on different datasets 
(NYU-V2 and VKITTI2).
We see that within each dataset, 
the weight distribution exhibits a fairly sharp peak.
The small bump on the right side of each plot is formed by 
the bias term in the weights
which reflects the average log depth of the dataset.

\subsection{Mid-density Depth Completion}

\begin{figure*}
\centering
\includegraphics[width=\linewidth]{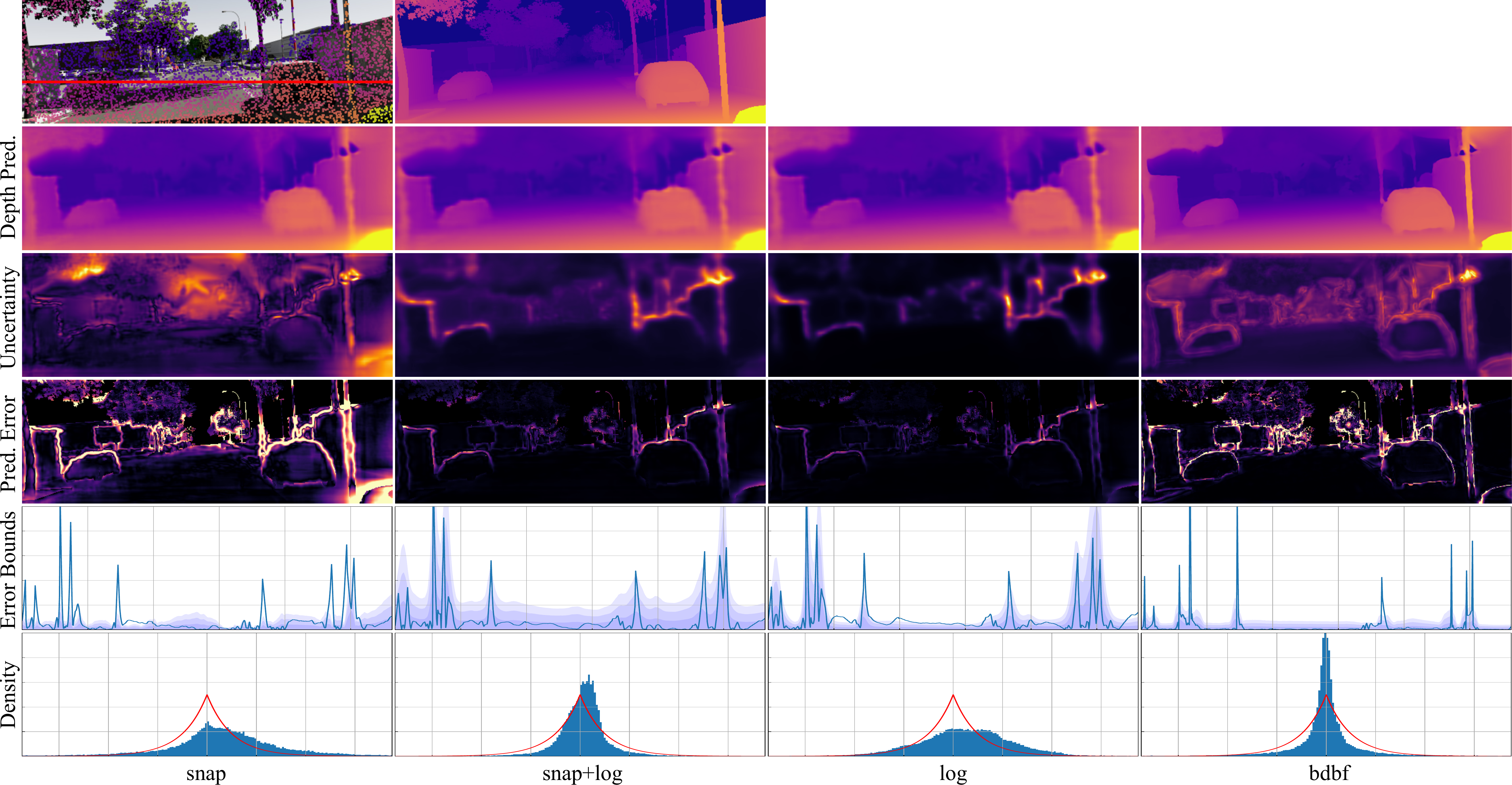}
\caption{
Sample qualitative results of all methods trained and test with 5\% sparsity
on VKITTI2.
Colormaps scales are different for each methods to visualize details.
Axes scales are same for all methods.
}
\label{fig:mid_vk2_i10}
\end{figure*}

\begin{figure*}
\centering
\includegraphics[width=\linewidth]{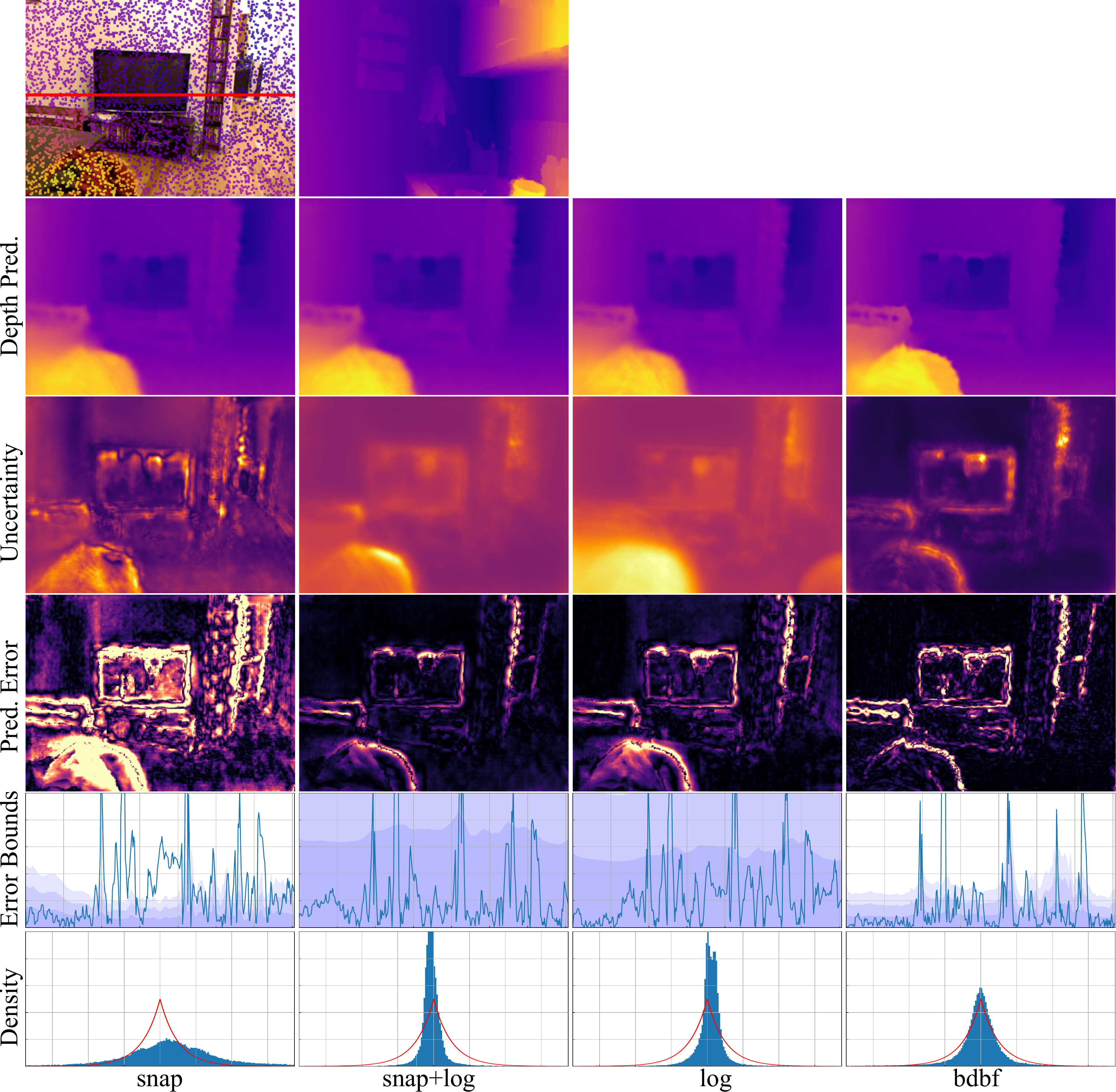}
\caption{
Sample qualitative results of all methods trained and test with 5\% sparsity
on NYU-V2.
Colormaps scales are different for each methods to visualize details.
Axes scales are same for all methods.
}
\label{fig:mid_nyu_i50}
\end{figure*}

\begin{table}
\centering\footnotesize
\begin{tabular}{|c|ccc|ccc|}
\hline
& \multicolumn{6}{c|}{Trained and tested on KITTI (original)}\\ \hline
Method   
& MAE   & RMSE  & $\delta_1$  
& AUSE  & AUCE  & NLL    \\ \hline

snap     
& 0.638 & \textbf{1.663} & 98.68
& 0.272 & 0.161 & -1.298 \\ 

snap+log
& 0.701 & 1.784 & 98.52
& 0.092 & 0.105 & -1.605 \\ 

log
& 0.899 & 2.022 & 98.44
& 0.109 & 0.133 & -1.544 \\ 

bdbf
& \textbf{0.364} & 1.682 & \textbf{98.74}
& \textbf{0.060} & \textbf{0.100} & \textbf{-2.337} \\ 

\hline\hline
& \multicolumn{6}{c|}{Trained on VKITTI2 and tested on KITTI (shifted)}\\ \hline
Method   
& MAE   & RMSE  & $\delta_1$  
& AUSE  & AUCE  & NLL    \\ \hline

snap     
& 1.296 & 2.549 & 92.03
& 0.876 & 0.161 & -0.675  \\ 

snap+log 
& 0.853 & 2.245 & 97.69
& 0.134 & 0.117 & -1.578  \\

log      
& 0.866 & 2.018 & 97.96 
& 0.138 & 0.274 & -0.755 \\

bdbf     
& \textbf{0.469} & \textbf{1.989} & \textbf{98.29}
& \textbf{0.072} & \textbf{0.037} & \textbf{-2.381} \\ \hline
\end{tabular}
\caption{
Quantitative results of all methods trained on KITTI and VKITTI  with 5\% sparsity and tested on KITTI.
}
\label{tab:mid_kitti}
\end{table}
Table \ref{tab:mid_5p_1p} shows quantitative results of 
all methods trained with 5\% sparsity and test on 5\% and 1\% sparsity.
Here we also include comparison with \textit{dbf}, 
where we simply ignore the prior during inference.
We see that \textit{dbf} and \textit{bdbf}
have perform similar due to the large amount of sparse depths,
but \textit{bdbf} still has the best performance overall.
Figure \ref{fig:mid_vk2_i10} and \ref{fig:mid_nyu_i50}
show some extra in-distribution samples of all methods 
from VKITTI2 and NYU-V2 respectively.

For domain shift, we take the same models that are trained on VKITTI2 
and directly test on KITTI without any fine-tuning. 
The data distribution shift in this case manifests most notably in image quality 
as well as noise level and spatial distribution of sparse depths.
Quantitative results are shown in Table~\ref{tab:mid_kitti}, 
where we also list in-distribution test on KITTI for comparison.
Most methods encounter performance drop in one or several metrics 
when tested under a slightly different domain.
However \textit{bdbf} achieve overall best results in both scenarios,
with its performance under distributional shift better then 
some of the baselines in the original domain.

For more tests on distributional shift, we utilize the 
\textbf{15-deg} and \textbf{30-deg} sequences from VKITTI2.
These sequences are variants of \textbf{clone} 
with the camera pointing at different angles.
Table \ref{tab:mid_vk2_ds} shows quantitative results of 
all methods trained with 5\% sparsity on VKITTI2 and 
test on these two sequences.

\subsection{Low-density Depth Completion}

Table \ref{tab:low_all_500_50_0} shows quantitative results of 
all methods trained with 500 sparse depths and test on 500 and 50 sparse depths.
We choose 50 because it is smaller than the number of basis (63).
This would previously fail with DBF, but BDBF have no problem dealing with any 
amount of sparse depths.
Moreover, we suggest that when designing a system working under extremely low sparsity
($\le$ 100 points)
it is better to forgo the idea of a depth encoder altogether, 
because convolution is simply not designed for sparse data
and interpolation schemes rarely work with only a few points.

\begin{figure*}
\centering
\includegraphics[width=\linewidth]{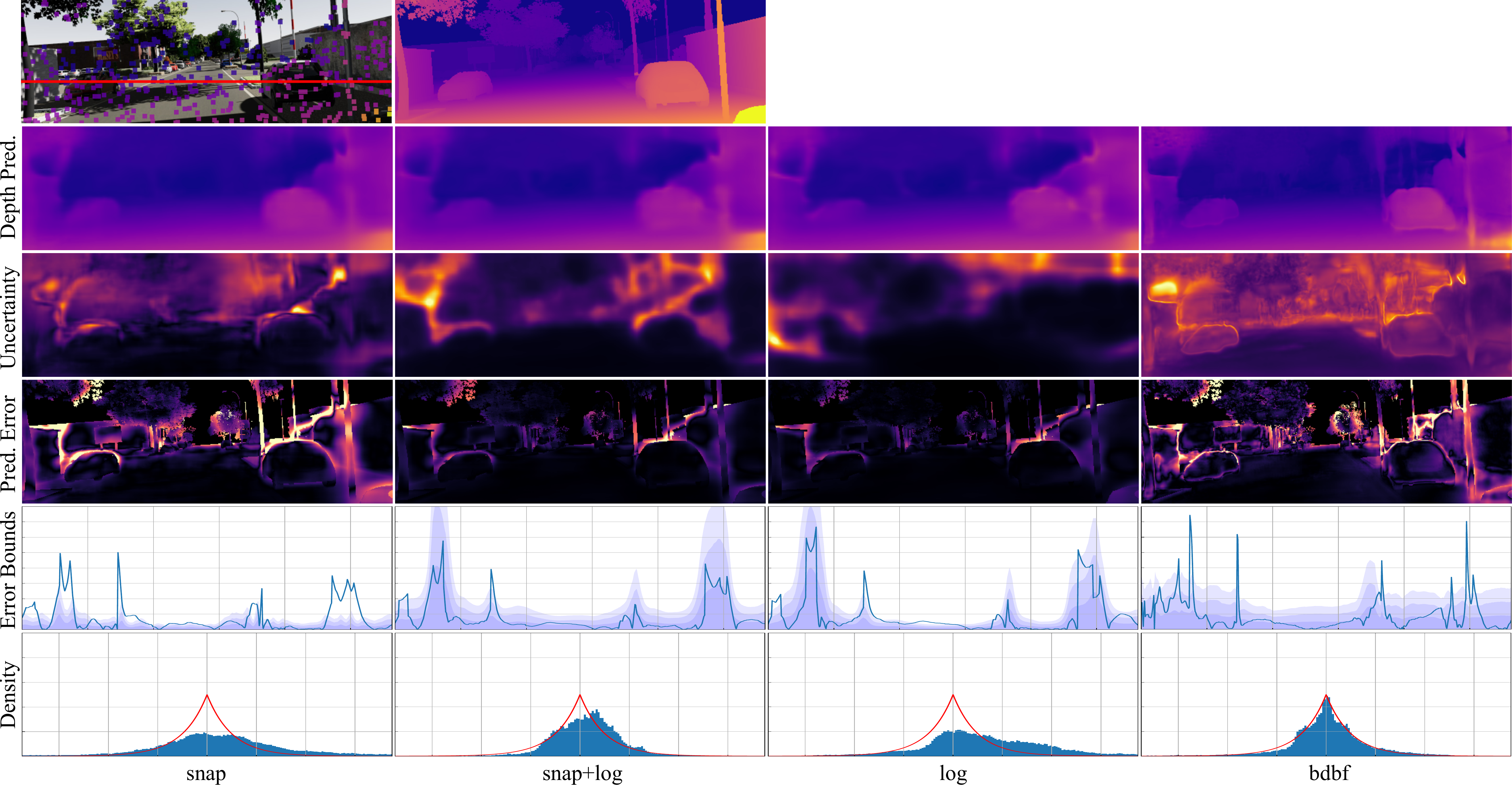}
\caption{
Sample qualitative results of all methods trained and test with 500 sparse points
on VKITTI2.
Colormaps scales are different for each methods to visualize details.
Axes scales are same for all methods.
}
\label{fig:low_vk2_i10}
\end{figure*}

\begin{figure*}
\centering
\includegraphics[width=\linewidth]{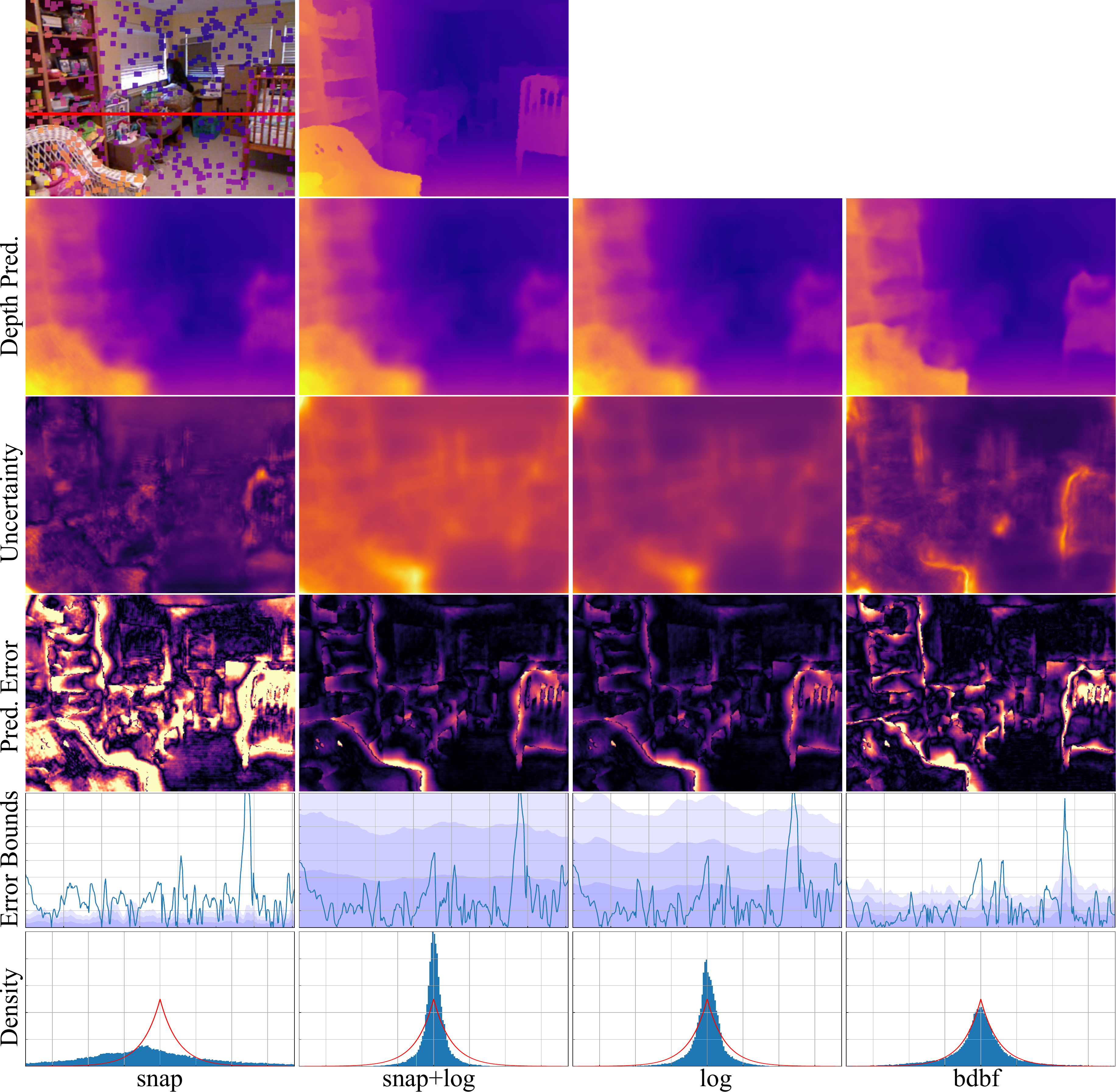}
\caption{
Sample qualitative results of all methods trained and test with 500 sparse points
on NYU-V2.
Colormaps scales are different for each methods to visualize details.
Axes scales are same for all methods.
}
\label{fig:low_nyu_i100}
\end{figure*}

Figure \ref{fig:low_vk2_i10} and \ref{fig:low_nyu_i100}
show some extra in-distribution samples of all methods 
from VKITTI2 and NYU-V2 respectively.

\begin{figure*}
\centering
\includegraphics[width=\linewidth]{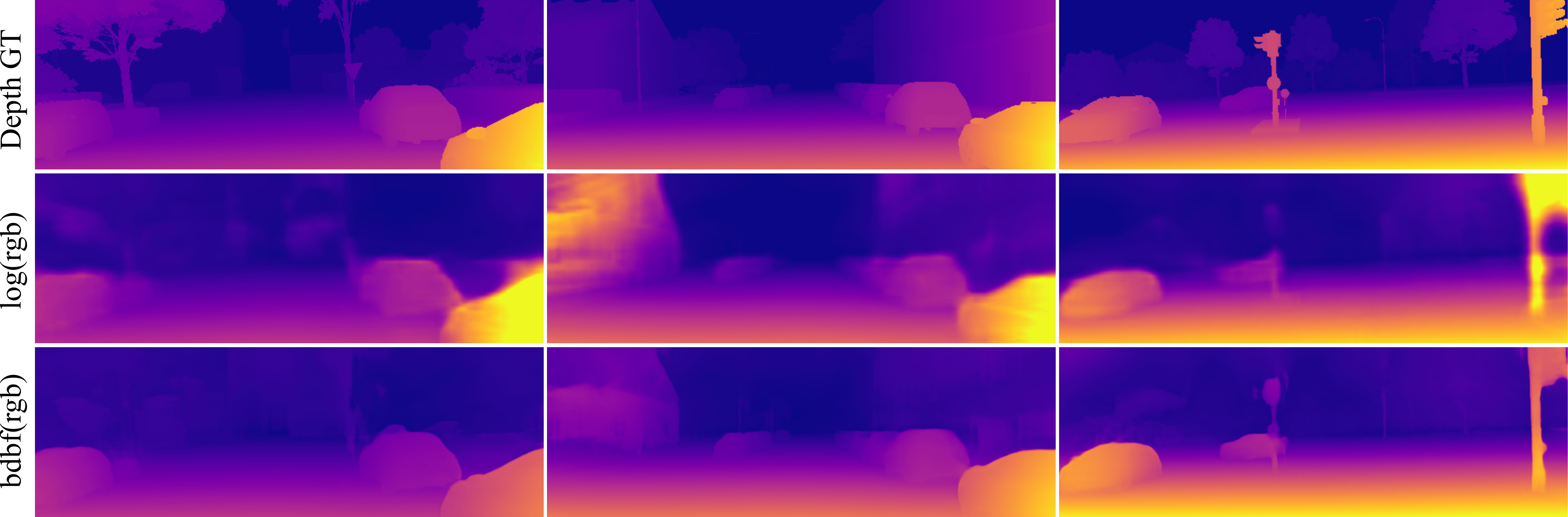}
\caption{
More qualitative results of our method tested with 0 sparse points.
\textit{log(rgb)} is trained as a monocular depth prediction network with NLL loss, 
which serves as a baseline.
\textit{bdbf(rgb)} is trained with 500 sparse depths.
}
\label{fig:low_vk2_n0}
\end{figure*}

Figure \ref{fig:low_vk2_n0} shows more qualitative results of 
\textit{bdbf(rgb)} and \textit{log(rgb)} on VKITTI2 with no sparse depths.

\begin{table}[t]
\centering\footnotesize
\begin{tabular}{|c|ccc|ccc|}
\hline

& \multicolumn{6}{c|}{VKITTI2 (15-deg)}\\ \hline
Method   
& MAE   & RMSE  & $\delta_1$  
& AUSE  & AUCE  & NLL    \\ 
\hline

snap     
& 1.164 & 3.161 & 95.67
& 0.445 & 0.172 & -0.733
\\ 

snap+log 
& 1.206 & 3.260 & 95.43
& 0.131 & \textbf{0.103} & -1.558 \\ 

log      
& 1.301 & 3.328 & 95.38
& 0.144 & 0.137 & -1.347 \\ 

bdbf     
& \textbf{0.686} & \textbf{2.861} & \textbf{97.92}
& \textbf{0.101} & 0.139 & \textbf{-2.618} \\ 
\hline
\hline

& \multicolumn{6}{c|}{VKITTI2 (30-deg)}\\ 
\hline
Method   
& MAE   & RMSE  & $\delta_1$  
& AUSE  & AUCE  & NLL \\ 
\hline

snap     
& 1.084 & 3.009 & 95.39
& 0.428 & 0.175 & -0.740 \\ 

snap+log 
& 1.139 & 3.106 & 95.10
& 0.128 & \textbf{0.087} & -1.509 \\ 

log      
& 1.217 & 3.166 & 95.10
& 0.141 & 0.146 & -1.251\\ 

bdbf     
& \textbf{0.627} & \textbf{2.711} & \textbf{98.02}
& \textbf{0.085} & 0.143 & \textbf{-2.676} \\ 
\hline
\end{tabular}
\caption{
Quantitative results of all methods trained with
5\% sparsity on VKITTI2 and test on 
\textbf{15-deg} and \textbf{30-deg} sequences from VKITTI2.
}
\label{tab:mid_vk2_ds}
\end{table}
\begin{table*}[t]
\centering
\footnotesize
\begin{tabular}{|c|c|c|ccc|ccc|ccc|ccc|}
\hline
\multicolumn{3}{|c|}{Trained with 5\%} 
& \multicolumn{6}{c|}{VKITTI2} 
& \multicolumn{6}{c|}{NYU-V2} \\ \hline 
Input & Method & \% 
& MAE & RMSE & $\delta_1$ & AUSE & AUCE & NLL 
& MAE & RMSE & $\delta_1$ & AUSE & AUCE & NLL
\\ \hline\hline

rgbd & snap & 5\% 
& 1.192 & 3.267 & 95.59 & 0.445 & 0.170 & -0.714
& 0.061 & 0.126 & 99.35 & 0.036 & 0.202 & -1.390
\\ 

rgbd & snap+log  & 5\% 
& 1.271 & 3.432 & 95.33 & 0.142 & \textbf{0.117} & -1.582
& 0.058 & 0.123 & 99.32 & 0.018 & 0.256 & -1.596
\\

rgbd & log & 5\%  
& 1.318 & 3.423  & 95.37 & 0.149 & 0.125 & -1.421
& 0.057 & 0.121  & 99.34 & 0.018 & 0.210  & -1.783
\\

rgbd & dbf & 5\%  
& 0.709 & 2.928  & 97.88 & 0.148 & 0.163 & -2.489
& 0.026 & 0.083  & 99.64 & 0.007 & 0.054  & -3.145
\\

rgbd & bdbf & 5\%  
& \textbf{0.703} & \textbf{2.925} & \textbf{97.88} & \textbf{0.110} & 0.136 & \textbf{-2.596} 
& \textbf{0.026} & \textbf{0.082} & \textbf{99.64} & \textbf{0.007} & \textbf{0.039} & \textbf{-3.151} 
\\ \hline 

rgbd & snap & 1\% 
& 1.784 & 4.679 & 92.04 & 0.805 & 0.214 & 0.730
& 0.087 & 0.198 & 97.65 & 0.051 & 0.238 & -0.387
\\ 

rgbd & snap+log  & 1\% 
& 1.805 & 4.755 & 92.03 & 0.235 & \textbf{0.062} & -1.333
& 0.089 & 0.211 & 97.35 & 0.025 & 0.202 & -1.451
\\

rgbd & log & 1\%  
& 1.831 & 4.769 & 92.21 & \textbf{0.223} & 0.147 & -1.168
& 0.087 & 0.214 & 97.31 & 0.023 & 0.157 & -1.605
\\

rgbd & dbf & 1\%  
& 1.393 & 4.384 & 94.52 & 0.350 & 0.110 & -1.670
& 0.055 & 0.155 & 98.45 & 0.016 & 0.072 & \textbf{-2.381}
\\

rgbd & bdbf & 1\%  
& \textbf{1.382} & \textbf{4.372} & \textbf{94.53}
& 0.271 & 0.082 & \textbf{-1.707}
& \textbf{0.055} & \textbf{0.155} & \textbf{98.45} 
& \textbf{0.016} & \textbf{0.059} & -2.374

\\ \hline 
\end{tabular}
\caption{
Quantitative results of all methods trained and tested with
5\% and 1\% sparsity on VKITTI2 and NYU-V2. 
}
\label{tab:mid_5p_1p}
\end{table*}
\begin{table*}[t]
\centering
\footnotesize
\begin{tabular}{|c|c|c|ccc|ccc|ccc|ccc|}
\hline
\multicolumn{3}{|c|}{Trained with 500} 
& \multicolumn{6}{c|}{VKITTI2} 
& \multicolumn{6}{c|}{NYU-V2} \\ \hline
Input & Method & \# 
& MAE & RMSE & $\delta_1$ & AUSE & AUCE & NLL 
& MAE & RMSE & $\delta_1$ & AUSE & AUCE & NLL 
\\ \hline\hline

rgbd & snap & 500 
& 2.312 & 5.403 & 90.14 & 0.459 & 0.229  & -0.207
& 0.096 & 0.206 & 97.53 & 0.053 & 0.261 & 0.211 
\\ 

rgbd & snap+log & 500 
& 2.396 & 5.571 & 89.88 & \textbf{0.273} & 0.036 & -1.150
& 0.095 & 0.213 & 97.44 & 0.025 & 0.205 & -1.393
\\

rgbd & log & 500 
& 2.492 & 5.800 & 89.13 & 0.299 & 0.095 & -0.906
& 0.097 & 0.212 & 97.44 & 0.025 & 0.152 & -1.502
\\ 

rgbd & dbf & 500 
& 2.050 & 5.067 & 92.58 & 0.453 & 0.051 & -1.175
& 0.065 & 0.168 & 98.45 & \textbf{0.020} & 0.055 & -2.186

\\ 
rgbd & bdbf & 500 
& \textbf{2.015} & \textbf{4.994} & \textbf{92.71} 
& 0.392 & \textbf{0.014} & \textbf{-1.215}
& \textbf{0.064} & \textbf{0.166} & 98.46 
& 0.021 & 0.030 & \textbf{-2.199} 
\\ 

\hline
rgb  & bdbf & 500 
& 2.569 & 5.642 & 88.67 & 0.481 & 0.015 & -0.979
& 0.098 & 0.199 & \textbf{98.48} & 0.030 & \textbf{0.014} & -1.689
\\ \hline \hline

rgbd  & snap & 50 
& 5.069 & 9.326 & 66.48 & 1.696 & 0.327 & 3.495
& 0.324 & 0.605 & 78.89 & 0.136 & 0.376 & 6.036
\\ 

rgbd  & snap+log & 50 
& 5.173 & 9.401 & 65.91 & 1.180 & 0.188 & 0.608
& 0.312 & 0.593 & 80.55 & 0.084 & 0.052 & -0.390
\\ 

rgbd  & log & 50 
& 5.107 & 9.433 & 66.52 & 1.114 & 0.275 & 2.439
& 0.323 & 0.599 & 80.49 & 0.088 & 0.134 & -0.166
\\ 

rgbd  & bdbf & 50 
& 4.098 & 8.440 & 76.91 & 0.977 & 0.021 & -0.411
& 0.215 & 0.446 & 88.63 & 0.081 & 0.026 & -0.926
\\ 

\hline
rgb  & bdbf & 50 
& \textbf{3.725} & \textbf{7.786} & \textbf{80.67}
& \textbf{0.701} & \textbf{0.012} & \textbf{-0.612}
& \textbf{0.144} & \textbf{0.296} & \textbf{94.97} 
& \textbf{0.039} & \textbf{0.011} & \textbf{-1.338}


\\ \hline
\end{tabular}
\caption{
Quantitative results of all methods trained and tested with 500 and 50 sparse depths.
\textit{rgbd} under the input column indicates the basis network uses the sparse depths 
scaffolding approach from~\cite{Wong2020UnsupervisedDC}, 
whereas \textit{rgb} uses color image as basis network input only.
}
\label{tab:low_all_500_50_0}
\end{table*}

{\small
\bibliographystyle{ieee_fullname}
\bibliography{reference}
}

\end{document}